\documentclass[final, 12pt]{elsarticle}
\usepackage{amsmath,graphicx, multirow}
\usepackage{breqn}
\usepackage{booktabs}
\usepackage{tabularx}
\usepackage{wrapfig, subcaption, setspace, booktabs}
\usepackage{xcolor}
\usepackage{paralist}
\usepackage{amsfonts}       
\usepackage{amsmath}
\usepackage{lscape}
\usepackage{makecell}
\usepackage{makecell}
\usepackage{boldline}
\usepackage{subscript}
\usepackage{tikz}
\usepackage{enumitem}
\usepackage{lineno}
\usepackage{soul}
\sethlcolor{green}

\usepackage{graphicx}
\usepackage{amssymb}

\journal{Pattern Recognition}

\def\checkmark{\tikz\fill[scale=0.4](0,.35) -- (.25,0) -- (1,.7) -- (.25,.15) -- cycle;}

\definecolor{mygold}{RGB}{165, 130, 114}
\definecolor{turcoise}{RGB}{164, 149, 141}
\definecolor{myh}{RGB}{138, 181, 92}

\begin{document}

\begin{frontmatter}


\title{Single-shot 3D multi-person pose estimation in complex images}



\author{Abdallah Benzine$^{\star,\dagger}$, Bertrand Luvison$^{\star}$, Quoc Cuong Pham$^{\star}$, Catherine Achard $^{\dagger}$}

\address{{\small $^{\star}$CEA LIST, Vision and Learning Lab for Scene Analysis, PC 184, F-91191 Gif-sur-Yvette, France} \\
   {\small $^{\dagger}$ Sorbonne University, CNRS, Institute for Intelligent Systems and Robotics, ISIR, F-75005 Paris, France }}

\begin{abstract}
In this paper, we propose a new single shot method for multi-person 3D human pose estimation in complex images. The model jointly learns to locate the human joints in the image, to estimate their 3D coordinates and to group these predictions into full human skeletons. The proposed method deals with a variable number of people and does not need bounding boxes to estimate the 3D poses. It leverages and extends the Stacked Hourglass Network and its multi-scale feature learning  to manage multi-person situations. Thus, we exploit a robust 3D human pose formulation to fully describe several 3D human poses even in case of strong occlusions or crops. Then, joint grouping and human pose estimation for an arbitrary number of people are performed using the associative embedding method. Our approach significantly outperforms the state of the art on the challenging CMU Panoptic and a previous single shot method on the MuPoTS-3D dataset. Furthermore, it leads to good results on the complex and synthetic images from the newly proposed JTA Dataset. 
\end{abstract}


\end{frontmatter}


\section{Introduction}
\label{sec:introduction}

3D human pose is a low dimensional and interpretable representation which is used a lot in action recognition \cite{presti20163d}.
3D human pose estimation based on RGB images is a challenging task from the computer vision perspective. Recent Convolution Neural Network (CNN) based approaches \cite{cao2016realtime, newell2017associative} achieve excellent performance in 2D human pose estimation thanks to large scale in the wild datasets.  Nevertheless, methods for 3D human pose estimation require 3D ground truth that is only available  using Motion Capture (Mocap) systems \cite{sigal2010humaneva, h36m_pami,mehta2016monocular}. Therefore, these methods have good performance in controlled environment but bad generalisation to real in the wild images. Furthermore, most of the 3D  pose  estimation  methods  are  restricted  to  a  single fully visible subject. In real-world scenarios, multiple people interact in cluttered or even crowded scenes containing both self-occlusions of the body and strong inter-person occlusions. Therefore, inferring the 3D pose of all the subjects (without knowing in advance their number) from a single and monocular RGB image is a harder problem and recent single-person 3D human pose estimation methods fail in this case.

A natural approach is to decompose the multi-person ill-posed problem into multiple single-person 3D estimations. These top-down approaches are based on the generation of multiple pose proposals that are evaluated and refined in a second time~\cite{rogez2017lcr}. Thus, they perform many redundant estimations and scale badly for a large number of subjects.

Another way to solve this problem is bottom-up strategy \cite{zanfir2018deep,zanfir2018monocular,mehta2017single}  that manages the whole scene in a single forward pass to give multi-person 3D human pose estimates. By their principle, they are more effective in managing occlusions between people and take advantage of context-related information to predict the different poses.

In the present article, we propose a new bottom-up approach that manages the whole scene in a single forward pass to give multi-person 3D human pose estimates. Our method is based on the Stacked Hourglass architecture~\cite{newell2016stacked} that has demonstrated its effectiveness for 2D human pose estimation. Single shot multi-person 3D human pose estimation is challenging as it needs to properly locate human joints and to regroup these estimations into final 3D skeletons. By  associating the Hourglass architecture with a powerful joints grouping method named the associative embedding \cite{newell2017associative}  and a robust multi-person 3D pose description \cite{mehta2017single}, we design an end-to-end architecture that jointly performs 2D human joints detection, joints grouping and full body 3D human pose estimation even when the subjects are partially occluded or truncated by the image boundary. The proposed method surpasses state of the art results on the CMU-Panoptic \cite{Joo_2017_TPAMI} dataset, achieves higher accuracy than a state of the art single-shot method on the MuPoTS-3D dataset \cite{mehta2017single}, and shows good results on  the Joint Track Auto dataset\cite{fabbri2018learning}, a synthetic but realistic dataset with a large number of people, various camera viewpoints and backgrounds. So far, this dataset has only been used for joint tracking. 

\section{Related Work}
\label{sec:related_work}

Human pose estimation is more and more studied as it is very useful for many applications (e.g. motion capture, human image synthesis, activity recognition, sign language recognition, robotics vision, etc.). In this section, we present recent deep learning approaches for 2D human pose estimation and single/multi-person 3D human pose estimation.\par
\vspace{0.2cm}
\textbf{2D human pose estimation}: Most methods for single-person 2D pose estimation extract probabilistic maps called heatmaps that estimates the probability of each pixel to contain a particular joint. At inference time, the 2D joint positions correspond to the local maxima of the heatmaps.  Most of these methods \cite{newell2016stacked, wei2016convolutional} are also iterative. A refined estimate of the heatmap is obtained from the previous estimates and the convolutional features. Wei \textit{et al.} \cite{wei2016convolutional} refine the predictions over successive stages with intermediate supervision at each stage. The Stacked Hourglass networks \cite{newell2016stacked} processes and consolidates features across scales to capture the spatial relationships of the human body.  \citet{bin2020structure} extend the Stacked Hourglass networks with a Pose Graph Convolutional Network to model the structural relationships between body key points. \citet{li2020exploring} introduce a Temporal Consistency Exploration module that captures geometric transformations between frames.

Both top-down and bottom-up human approaches have been proposed for multi-person 2D human pose estimation. Top down methods \cite{papandreou2017towards, he2017mask} first detect human bounding boxes and then estimate 2D human poses. Nevertheless, these methods fail when the detector fails, in particular when there are strong occlusions. Bottom-up approaches \cite{cao2016realtime,newell2017associative} first estimate the 2D location of each joint and then associate them into full skeletons. \citet{cao2016realtime} regress affinity between joints that means the direction of the bones in the image.  Unlike this approach that needs complex post-processing  joints,  \citet{newell2017associative} propose to learn this association in an end-to-end network thanks to the Associative Embeddings. \citet{zhao2020cluster} exploit multi-level contextual association with a cluster-wise feature aggregation network.
\par
\vspace{0.2cm}
\textbf{Single-person 3D human pose estimation}: Motivated by the recent advances in 2D human pose estimation, some existing approaches \cite{martinez2017simple, fang2018learning, bogo2016keep, simo2012single, wang2014robust, ramakrishna2012reconstructing, chen20173d, moreno20173d, nie2017monocular, atrevi2017very} use only 2D human poses estimated by other methods \cite{newell2016stacked, cao2016realtime} to predict 3D human poses. \citet{atrevi2017very} perform
2D body silhouette matching to assign 3D joints.
\citet{chen20173d} performs a nearest neighbour search on a given 3D pose library with a large number of 2D projections.  \citet{moreno20173d} formulate the problem of the 3D human pose estimation as a 2D to 3D distance matrix regression. Nie \textit{et al.} \cite{nie2017monocular} predict depth on joints using LSTM. Martinez \textit{et al.} \cite{martinez2017simple} lift 2D joints to 3D space using a deep residual neural network. Nevertheless, these approaches are limited by the 2D pose estimator performance and do not take into account important images clues, such as contextual information, to make the prediction.

Other methods predict 3D human poses from images features\cite{agarwal20043d, rogez2008randomized, sminchisescu2004generative, bo2008fast, shakhnarovich2003fast}. Recent methods make this prediction directly from  monocular images \cite{ pavlakos2017coarse, mehta2017vnect, popa2017deep, chen2016synthesizing, rogez2016mocap, li20143d,madadi2020smplr} or from sequences of images \cite{tekin2016direct, zhou2016sparseness} using Convolutional Neural Networks. The learning procedure needs images annotated with 3D ground-truth pose. Since no large scale 3D in the wild annotated dataset exists,  current approaches tend to overfeat on the constrained environment they have been trained on.  The existing in the wild approaches use either synthetic data \cite{ chen2016synthesizing, rogez2016mocap,varol2017learning} or are trained on both 3D and  in the wild 2D datasets \cite{mehta2017monocular,sun2017compositional,  simo2013joint, zhou2014spatio, tekin2017learning,  zhou2017towards, pavlakos2018ordinal, yang20183d}.  Mehta \textit{et al.} \cite{mehta2017monocular} use a pretrained 2D pose network to initialize the 3D pose regression network.  Zhou \textit{et al.}  use geometric constraints \cite{zhou2017towards} in a weakly supervised setting. Pavlakos \textit{et al.} \cite{pavlakos2018ordinal} take another approach by relying on weak 3D supervision
in form of a relative 3D ordering of joints which can be easily annotated even for in the wild images. Yang \textit{et al.} \cite{yang20183d} use an adversarial loss that transfers the 3D human pose structures
learned from the indoor annotated dataset to the in-the-wild images. Although performing well with a single fully visible subject, these methods fail with several interacting people that are at different image scale and that occult each other.   \par
\vspace{0.2cm}
\textbf{Multi-person 3D human pose estimation}: In a top-down approach, Rogez \textit{et al.} \cite{rogez2017lcr, rogez2019lcr} generate human pose proposals that are further refined using a regressor. Moon et al. \cite{moon2019camera} propose a camera distance aware multi-person top-down approach that performs human detection (DetectNet), absolute 3D
human localisation (RootNet) and root relative 3D human pose estimation (PoseNet) for each person independently. Zanfir \textit{et al.} \cite{zanfir2018monocular} estimate the 3D human shape from sequences of frames using a pipeline process followed by a 3D pose refinement based on a non-linear optimisation process and semantic constraints.   MubyNet \cite{zanfir2018deep} is a bottom-up multi-task network that identifies joints and learns to score their possible associations as limbs. These scores are used to solve a global optimisation problem that groups the joints into full skeletons following the human kinematic tree.  
Mehta \textit{et al.} \cite{mehta2017single} propose an approach that predicts 2D heatmaps, part affinity fields \cite{cao2016realtime} and Occlusions Robust Pose Maps (ORPM). This approach manages multi-person 3D human pose estimation even for occluded and cropped people.  Nevertheless, the architecture used in \cite{mehta2017single} is not a stacked architecture while the stacking strategy \cite{cao2016realtime, newell2017associative} performs well in the 2D context. 

The proposed method deals with multi-person 3D human pose estimation. Unlike \cite{zanfir2018monocular}, it does not need sequence of images to refine the pose estimates. It is based on the stacked hourglass networks \cite{newell2016stacked} devoted to mono-person 2D pose estimation and showing very good performance on this task. Thus, we extend this approach using the multi-person 3D poses description robust to occlusions proposed in \cite{mehta2017single} and the  associative embedding \cite{newell2017associative} to group joints into full skeletons. The final network architecture is notably trained in an end-to-end manner and the inference requires a single forward pass.  Our work is similar to \citet{mehta2017single} as both methods perform bottom-up 3D multi-person pose estimation but differ in two ways. First, a stacked architecture is used while a ResNet-50 \cite{he2016deep} is used in \cite{mehta2017single}. Recent works show the effectiveness of such a refinement strategy for 2D pose estimation \cite{cao2016realtime, newell2016stacked} but also for 3D pose estimation \cite{zhou2017towards}. Secondly, our work differ in the grouping method used to group joints' detections into full human skeletons. Part Affinity Fields are used in \cite{mehta2017single} which may be a sub-optimal way of grouping joints because the grouping is performed by solving a bipartite graph matching problem while the associative embedding method is a more direct way to perform this grouping. Indeed, no multi-stage pipelines is required in our model and the network simultaneously learn to perform pose estimation and joints' grouping. Furthermore, the experimental results detailed in \citet{newell2017associative} show that the associative embedding method is more effective than Part Affinity Fields in a 2D context.

\section{Proposed Method}
\label{sec:proposed_approach}

\begin{figure*}[htb]

\begin{minipage}[ht]{\textwidth}
\begin{minipage}[t]{0.45\textwidth}
  \centering
  \centerline{\includegraphics[width=\textwidth]{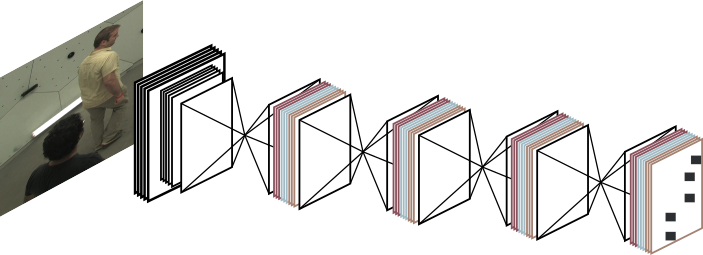}}
  \centerline{(a) Stacked Hourglass architecture}\medskip
\end{minipage}
\hfill
\begin{minipage}[t]{0.45\textwidth}
  \centering
  \centerline{\includegraphics[width=\textwidth]{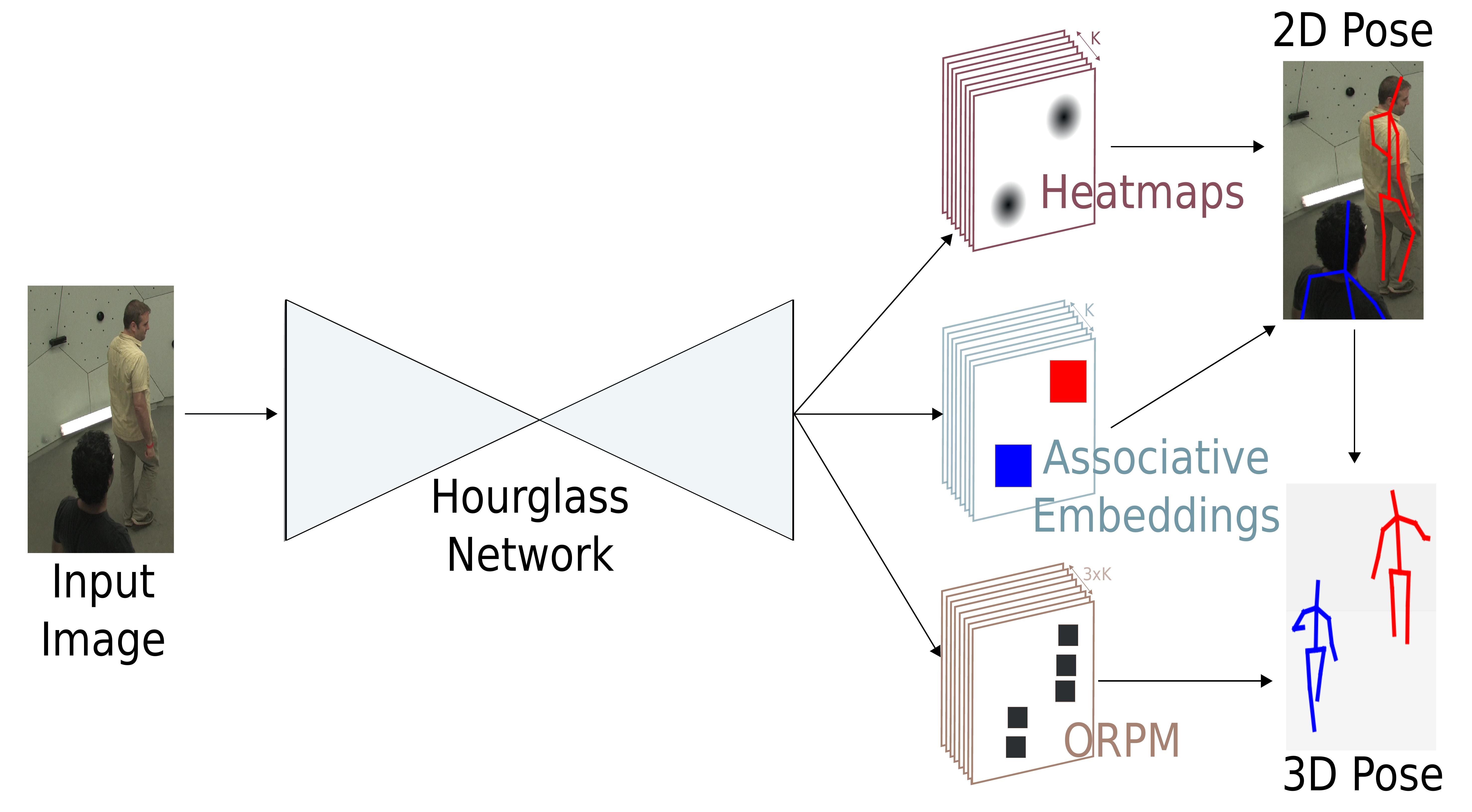}}
  \centerline{(b) Hourglass Network predictions}\medskip
\end{minipage}
\hfill
\begin{minipage}[ht]{\textwidth}
  \centering
  \centerline{\includegraphics[width=\textwidth]{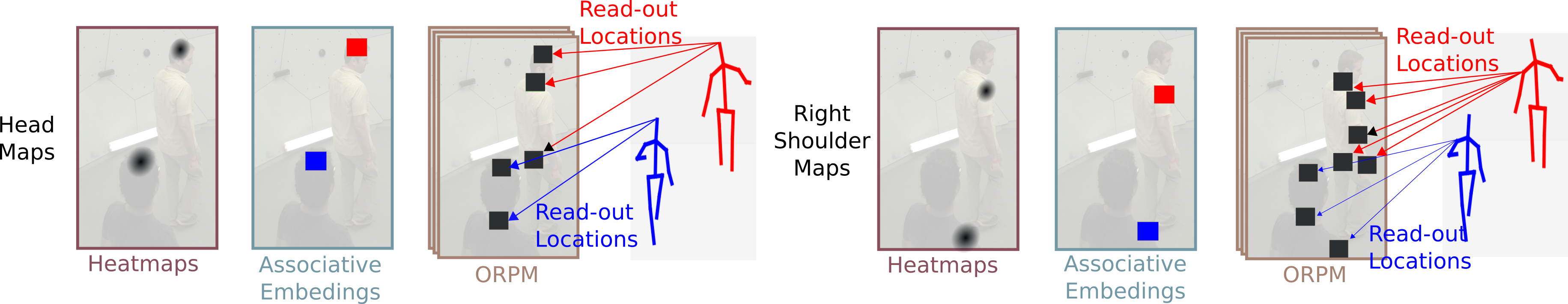}}
  \centerline{(c) Predicted Maps}\medskip
\end{minipage}
\end{minipage}

\caption{The proposed model estimates full 3D skeletons for an arbitrary number of people. The stacked hourglass architecture, depicted in a), is used. Figure b) illustrates the prediction of each hourglass network that predicts, for each joint, a \textcolor{myh}{2D localisation map (heatmap)}, an \textcolor{turcoise}{associative embedding map} and \textcolor{mygold}{Occlusion Robust Pose Maps (ORPM)}. These maps are successively refined by the following hourglass module. The last hourglass produces the final result.  
Figure c) illustrates in more detail the maps predicted by an hourglass network. The associative embedding maps contain different embedding values for joints belonging to different subjects and close embedding for joints belonging to the same subject. The ORPM store 3D joints coordinates at different 2D locations. All joints' coordinates are stored at the position of two root joints (neck and pelvis) at the corresponding OPRM maps. These coordinates are also stored at the 2D positions of the joint's limb.  For instance, the shoulder 3D coordinates are stored in the ORPM at the shoulder 2D position but also at the elbow and wrist positions. This redundancy allows 3D full pose readout even in case of strong occlusions and cropping. Two examples are presented for the head (left) and the right shoudler (right). Best viewed in color.}
\label{fig:overview}
\end{figure*}

\subsection{Description}

Given a monocular RGB image \textbf{I} of size $\mathrm W \times \mathrm H$, we seek to estimate the 3D human poses $\mathcal P_I=\{P_i\, |\, i \in [1,\ldots, N] \} $  where $N$ is the number of visible people,  $P_i \in \mathbb{R}^{3 \times K}$ are the 3D joints locations and $K$ is the number of predicted joints.  The 3D joint  coordinates are expressed relatively to their parents joints in the kinematic tree and converted to pelvis relative locations for evaluation in a 3D coordinate reference oriented like the camera one. The model is composed of several stacked hourglass networks. The image is first sub-sampled to images features \textbf{I'} of size $W' \times H'$ by convolution and pooling layers.  Each hourglass module outputs heatmaps for 2D joints detection, ORPM for 3D joints localisation and associative embeddings maps for joint grouping, each map being of size $W' \times H'$. Except for the first hourglass that takes as input only image features \textbf{I'}, other hourglasses takes as input images features \textbf{I'} and the prediction of the previous hourglass that is refined. Figure 1~\ref{fig:overview} depicts an overview of the proposed method. 
\vspace{-0.1cm}
\subsection{Occlusions Robust Pose Maps}
\label{sec:orpm}

Suppose we have an image I and the corresponding 3D poses $P_I$. A good 3D pose representation to train a Convolutional Neural Network should have the following characteristics: 
\begin{itemize}
    \item a fixed dimension regardless of the number of people in the image;
    \item being robust to occlusions and crops. 
\end{itemize}

To address these two problems, we adopt the ORPM formulation. For each joint, each hourglass network outputs three maps of dimensions $W'\times H'$, one for each X,Y,Z dimension. The size of these maps does not depend on the number of visible people which allows the estimation of the 3D pose of an arbitrary number of people. In these maps, the 3D joint coordinates of each person are stored at different 2D locations: 
\begin{itemize}
    \item at the 2D positions of the pelvis and the neck;
    \item at the 2D position of the joint;
    \item at the 2D positions of the joints belonging to the same limb.
\end{itemize}
For instance, the 3D coordinates of the wrist joints are stored in the wrist ORPM at the pelvis, the neck, the elbow and the shoulder 2D positions.  This redundancy in the ORPM allows a robustness to occlusions and crops. Indeed, neck and pelvis are the joints that are the best estimated and the less prone to occlusions. \par 
At inference time, the 3D pose readout of a person is performed in two steps: a full 3D pose readout followed by a 3D pose refinement. 
\par
The full 3D pose readout is performed by reading the full person 3D pose at the following 2D positions in the ORPM: 
\begin{itemize}
    \item at the pelvis 2D position, if the pelvis is detected;
    \item at the neck 2D position,  if the neck is detected and the pelvis is not.
\end{itemize}


The full 3D pose readout is followed by the 3D pose refinement. During this step, for each joint, we refine the predicted 3D coordinates previously obtained by reading in the ORPM at one of the following 2D locations: 
\begin{itemize}
    \item at the joint 2D position in the ORPM if this 2D position is a valid readout location;
    \item at the 2D position of a joint belonging to the joint's limb.  We take the extremity of the joint's limb and we go back in the kinematic tree until a valid readout location is found.
\end{itemize}

If no valid readout location is found in the joint's limb, the 3D coordinates are not refined. A 2D readout position is considered valid if it satisfies the following criteria: 
\begin{itemize}
    \item the confidence associated to the 2D predicted position of the joint is higher than a given threshold $\tau_C$;
    \item  the distance between the 2D joint position and the 2D position of the other joints must be less than a given distance $\tau_D$;
    \item the 3D coordinates read at this 2D position in the ORPM must be anthropomorphically correct. In this purpose, we compute the mean length of each limb in the training dataset and we reject each predicted 3D coordinates that gives limbs whose length is too far from the corresponding computed mean. 
\end{itemize}

\subsection{Associative embedding}

The network predicts for each joint a 2D heatmap and 3 ORMP for each $X, Y, Z$ joint coordinates.  This description is independent of the number of people. Now, we use the associative embedding to associate the joint to full skeletons.  
Predicted heatmaps contain peaks at the 2D joint positions of different subjects. To regroup the joints belonging to the same person, an additional output is added to the network for each joint corresponding to embeddings.  Detections are then grouped by comparing the embedding values of different joints at each 2D peak position in the heatmap. If two joints have a close embedding value, they belong to the same person. The network is trained to perform this grouping by predicting close embeddings for joints belonging to the same person and distant embeddings for joints of distinct people.

Formally, let $E_k \in \mathbb{R}^{W'\times H'}$ be an embedding map predicted by the network for the $k^{\text{th}}$ joint and $e_k(\boldsymbol{x})$ be the embedding value at the 2D position $\textbf{x}$. Let us consider an image composed of $N$ people, each having $K$ joints. Let $\boldsymbol{x_{k,n}}$ be the 2D ground-truth position of the $k^{\text{th}}$ joint of the person $n$. We refer by \textit{reference embedding}, the predicted embedding of a person obtained as the mean of its embedding's joints:
\begin{equation}
 \overline{e}_n = \frac{1}{K} \sum_{k}e_k(\boldsymbol{x_{k,n}})
\end{equation}

The grouping loss is then defined by:

\begin{equation}
 \mathcal{L}_{AE}= \frac{1}{NK}\sum_n \sum_k \left( \overline{e}_n - e_k(x_{k,n}) \right)^2 + \frac{1}{N^2} \sum_n \sum_{n'\neq n} \exp \left( - \frac{1}{2 \sigma^2} ( \overline{e}_n -  \overline{e}_{n'} )^2 \right)
\label{equation_ae}
\end{equation}

The first term of equation \eqref{equation_ae} corresponds to a pull loss that brings similar embeddings for  joints belonging to a same person and the second part corresponds to a push loss that gives different embeddings to  joints of different subjects. $\sigma$ is a parameter giving more or less importance to the push loss. It has been experimentally fixed to 1. 

\vspace{-0.2cm}
\subsection{Network loss}

We learn jointly the three following tasks: 
\begin{inparaenum}[i)]
   \item 2D joint localisation by predicting heatmaps;
    \item 3D joint coordinates estimation with ORPM prediction;
    \item Joint grouping with associative embedding prediction.
   \end{inparaenum} The network loss is then: 
\begin{equation}
    \mathcal{L}_{\text{3DMP}} =    \mathcal{L}_{\text{2D}} + \, \mathcal{L}_{\text{ORPM}} + \lambda_{\text{AE}}\, \mathcal{L}_{\text{AE}}
\end{equation}

Where $\mathcal{L}_{\text{2D}}$ is the euclidean distance between the ground-truth 2D heatmaps and the predicted 2D heatmaps, $\mathcal{L}_{\text{ORPM}}$ is the euclidean distance between the predicted ORMP and the ground-truth ORMP and $\mathcal{L}_{\text{AE}}$ is the loss defined by equation \eqref{equation_ae}. $\lambda_{\text{AE}}=0.001$ is the weight of the Associative Embeddings loss.

\subsection{Multi-Scale Inference}
\label{multi_scale_infrence}

\begin{figure*}[htb]

  \centerline{\includegraphics[width=\textwidth]{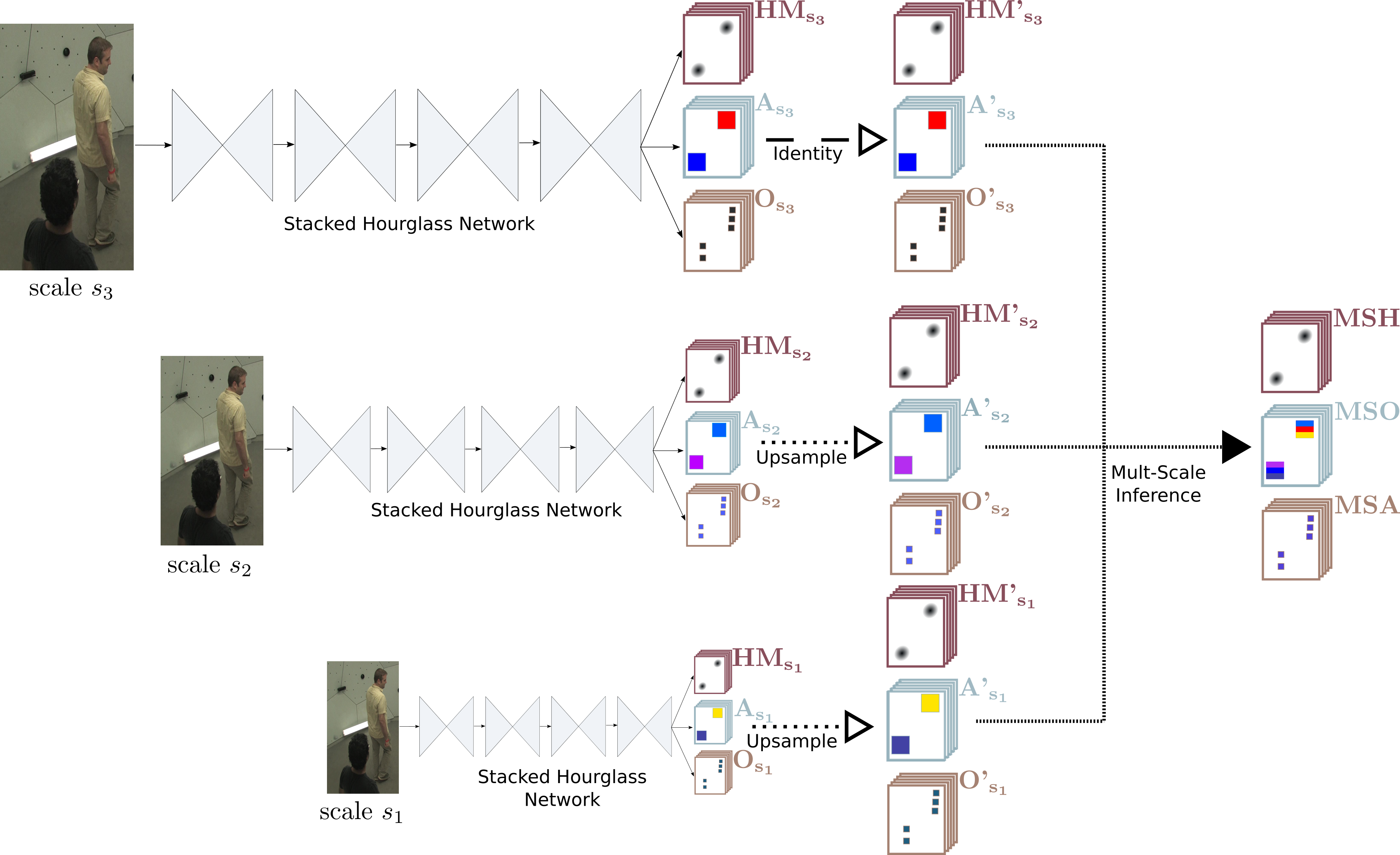}}

\caption{\textbf{Overview of the Multi-Scale Inference(MSI)}. MSI combines the predictions made by the network at different image scales allowing the  3D poses prediction of people with different resolution. For each scale $s_i$, heatmaps $\textbf{HM\textsubscript{s\textsubscript{i}}}$, ORPM $\textbf{O\textsubscript{s\textsubscript{i}}}$ and Associative Embedding maps  $\textbf{A\textsubscript{s\textsubscript{i}} }$ are predicted. These maps are then resized to the highest scale ($s_3$ here) and then combined following the procedure described in subsection \ref{multi_scale_infrence} to obtain the final Multi-Scale Heatmaps (MSH), Multi-Scale Associative embedings (MSA) and  Multi-Scale ORPM (MSO). Best viewed in color.}
\label{fig:MSI}
\end{figure*}

Although being single-shot and working well when there are a reduced number of people that are close to the camera, our method with a single scale inference tends to fail in complex and crowded images like those from the JTA dataset. In these images, visible people are at very different distances from the camera.  Consequently, these people are projected with very different pixel resolutions and the model has difficulties to handle properly all these scales with a single image resolution. To handle these cases, Multi-Scale Heatmaps, Multi-Scale Associative Embeddings  maps and Multi-Scale ORPM are computed. 

Suppose that we have an input image \textbf{I} for which we want to extract Multi-Scale Heatmaps, Multi-Scale Associative Embeddings maps and Multi-Scale ORPM.  Let $\textbf{\textrm{S}}={s_1, s_2, \dots , s_M}$ the scale pyramid for which we want to compute these maps, $s_M$ being the highest resolution scale. 

First, for each scale $s_i$, we compute $\textbf{HM\textsubscript{s\textsubscript{i}}} \in  \mathbb{R}^{K \times W_{s_i}\times H_{s_i}} $ , $\textbf{A\textsubscript{s\textsubscript{i}} }\in  \mathbb{R}^{K \times W_{s_i}\times H{s_i}} $ , $\textbf{O\textsubscript{s\textsubscript{i}}} \in  \mathbb{R}^{3 \times K \times W_{s_i}\times H{s_i}} $ respectively the predicted heatmaps, associative embedding maps and ORPM for scale $s_i$. Each $\textbf{HM\textsubscript{s\textsubscript{i}}}$,   $\textbf{A\textsubscript{s\textsubscript{i}} }$ and  $\textbf{O\textsubscript{s\textsubscript{i}}}$ is resized to maps $\textbf{HM'\textsubscript{s\textsubscript{i}}}$,   $\textbf{A'\textsubscript{s\textsubscript{i}} }$ and  $\textbf{O'\textsubscript{s\textsubscript{i}}}$ that match the resolution of scale $s_M$, as shown in Figure \ref{fig:MSI}. 

The Multi-Scale Heatmaps are the mean of the rescaled heatmaps. Let $ \textbf{MSH} \in  \mathbb{R}^{K \times W_{s_M}\times H_{s_M}} $ be the Multi-Scale Heatmaps, $msh(j,\boldsymbol{x})$ be the value of $\textbf{MSH}$ at position $\boldsymbol{x}$ for joint $j$ and $h'_{s_i}(j,\boldsymbol{x})$ be this value for the rescaled heatmaps $\textbf{HM'\textsubscript{s\textsubscript{i}}}$. Then, we have : 
\begin{equation}
msh(j,\boldsymbol{x})  = \frac{1}{M} \sum_{i=1}^{M}h'_{s_i}(j,\boldsymbol{x})
\end{equation}

The Multi-Sacle Associative Embeddings maps $\textbf{MSA} \in  \mathbb{R}^{K \times W_{s_M}\times H_{s_M} \times s_M }$  are the concatenation of the rescaled associative emedings maps $\textbf{A'\textsubscript{s\textsubscript{i}} }$.

In order to compute the Multi-Scale ORMP $ \textbf{\textrm{MSO}} \in  \mathbb{R}^{3\times K \times W_{s_M}\times H_{s_M}} $, we cannot  compute a simple average like done for the heatmaps. Indeed, if a person is detected at a given scale but not in another one, if we simply compute the average between the ORPM at each scale, the well estimated 3D pose at one scale could be altered by this operation. To avoid this, the mean is weighted by the predicted heatmaps and we take into account the different readout locations induced by the ORPM formulation.  This way, more the model is confident about a predicted joint at a given scale, more the ORPM at this scale will contribute to the $\textbf{MSO}$. Let $mso(c, j,\boldsymbol{x})$ be the value of $\textbf{\textrm{MSO}}$ for coordinate $c$ (X, Y or Z coordinate) and joint $j$ at 2D position $\boldsymbol{x}$, $O'_{s_i}(c,j,\boldsymbol{x})$ be this value for the rescaled ORPM $\textbf{O'\textsubscript{s\textsubscript{i}}}$  and $RL^j= {rl_1^j, rl_2^j, \dots, rl_{L_j}^j}$ be the set of readout locations for the joint j  in the ORPM. Then, we have: 
\begin{equation}
mso(c,j,\boldsymbol{x}) = \frac{\sum_{i=1}^M\sum_{l=1}^{L_j} h_{s_i}'(rl^j_l, \boldsymbol{x})o_{s_i}'(c,j,\boldsymbol{x})}{\sum_{i=1}^M\sum_{l=1}^{L_j} h_{s_i}'(rl^j_l, \boldsymbol{x})}
\end{equation}

\vspace{-0.2cm}
\subsection{Final prediction}

Once the network is trained, the final prediction is obtained in several stages. First, a non-maximum suppression is applied on the heatmaps to obtain the set of joint detections. Then, all the neck embeddings are read from the neck embedding map at the predicted neck 2D positions. This pool of 2D neck positions with their corresponding embedding gives the initial set of detected people. The other joints associated to these necks need now to be found. Each person is characterised by its reference embedding. The next joint associated to a given person is the one having the highest detection score and having a distance with the person embedding lower than a given threshold $\tau_{AE}$. We repeat this step until there is no more joint that respects these two criteria. Once this process is done, the  non-associated joints are used to create a new pool of people. At the end, the 2D pose of each person is obtained and used to read the 3D pose in the ORPM as described in Section~\ref{sec:orpm}.

\section{Experiments}

In this paper, we address the problem of single shot multi-person 3D human pose estimation. To evaluate our method, we perform separate experiments on: 
\begin{itemize}
    \item single-person 3D pose estimation in a controlled environment (Human 3.6M dataset \cite{ionescu2014human3})
    \item multi-person 3D pose estimation in a controlled environment (CMU-Panoptic dataset \cite{Joo_2017_TPAMI}); some images are depicted in Figure \ref{fig:res_panoptic}.
    \item multi-person 3D pose estimation in outdoor and indoor scenes (MuPoTS-3D dataset \cite{mehta2017single}).
    \item multi-person 3D pose estimation in virtual environments with many people (JTA dataset \cite{fabbri2018learning}). This dataset is more complex and richer than the previous one. Some images are shown in Figure \ref{fig:res_jta}. No previous method for 3D human pose estimation has been evaluated on this dataset to the best of our knowledge. 
\end{itemize}

\textbf{Evaluation Metrics: } To evaluate our Multi-Person 3D pose approach, we use two metrics. The first one is the Mean per Joint Position Error (MPJPE) that corresponds to mean Euclidean distance between ground truth and prediction for all people and all joint. The second one is the 3DPCK which is the 3D extension of the Percentage of Correct Keypoints (PCK) metric used for 2D Pose evaluation, as well. A joint is considered correctly estimated if the error in its estimation is less than 150mm. If an annotated subject is not detected by our approach, we consider all of its joints to be incorrect in the 3DPCK metric.  
We distinguish between $3DPCK_r$ that is calculated after root joints alignment and $3DPCK_a$ that is calculated in the orginal camera 3D space.

\textbf{Training Procedure}: The method was implemented with PyTorch. The hourglass component is based on the public code in \cite{newell2017associative}. We used four stacked hourglasses in our model, each one outputting 2D heatmaps, ORPM and associative embeddings. We trained the model using mini-batches of size 30 on 8 Nvidia Titan X GPU during 240k iterations. We used the Adam\cite{kingma2014adam} optimiser with an initial learning rate of $10^{-4}$.

\subsection{Single-person 3D pose estimation on Human 3.6M}
Human 3.6M \cite{ionescu2014human3} is a dataset containing 3.6 million single-person RGB images with 3D human poses annotated by MoCap systems. We used the standard protocol for the evaluation: S1, S5, S6, S7 and S8 subjects for training and the subjects S9 and S11 for testing.

\begin{table}[!t]
\centering
\resizebox{\linewidth}{!}{%
\begin{tabular}{@{}lllllllll@{}}
\toprule
                     & Direction & Discussion & Eating & Greet & Phone & Photo & Pose  & Purchase \\ \midrule

\cite{pavlakos2017coarse} & 67.4      & 71.9       & 66.7   & 69.1  & 72.0  & 77.0  & 65.0  & 68.3     \\

\cite{martinez2017simple} & 51.8 & 56.2 & 58.1 & 59.0 & 69.5 & 78.4 & 55.2 & 58.1 \\
\cite{mehta2017monocular} & 52.5 & 63.8 & \textbf{55.4} & 62.3 & 71.8 & 79.8 & \textbf{52.6} & 72.2 \\
\cite{mehta2017vnect} & 62.6 & 78.1 & 63.4 & 72.5 & 88.3 & 93.8 & 63.1 & 74.8 \\ 
\cite{zhou2017towards}          & 54.8      & 60.7       & 58.2   & 71.4  & \textbf{62.0}  & \textbf{65.5}  & 53.8  & 55.6     \\
\cite{rogez2017lcr} & 76.2 & 80.2 & 75.8 & 83.3 & 92.2 & 79.9 & 105.7 &  71.7 \\
\cite{mehta2017single} & 58.2 & 67.3 & 61.2 & 65.7 & 75.82 & 84.5 & 62.2 & 64.6 \\ 
\cite{fang2018learning}     & \textbf{50.1}      & \textbf{54.3}       & 57.0   & \textbf{57.1}  & 66.6  & 73.3  & 53.4  & 55.7     \\
\midrule

Ours                 & \textbf{50.1 }     & 66.4       & 56.4   & 65.0  & 69.4 & 81.5  & 55.6  & \textbf{52.1 }    \\ \toprule
                     & Sitting   & SittingD   & Smoke  & Wait  & WalkD & Walk  & WalkT & AVG      \\\midrule

\cite{pavlakos2017coarse} & 83.7      & 96.5       & 71.7   & 65.8  & 74.9  & 59.1  & 63.2  & 71.9     \\


\cite{martinez2017simple} & 74.0 & 94.6 & 62.3 & 59.1 & 65.1 & 49.5 & 52.4 & 62.9 \\
\cite{mehta2017monocular} &  86.2 & 120.6 & 66.0 & 64.0 & 76.8  & 48.9 & 53.7 & 68.6 \\
\cite{mehta2017vnect} & 106.6 & 138.7 & 93.8 & 73.9 &82.0  & 55.8 & 59.6 & 80.5 \\
\cite{zhou2017towards}          & 75.2      & 111.6      & 64.2   & 66.1  & \textbf{51.4}  & 63.2  & 55.3  & 64.9     \\ 
\cite{rogez2017lcr} & 105.9 & 127.1 & 88.0 & 83.7 & 86.6  & 64.9& 84.0 & 87.7 \\
\cite{mehta2017single} & 82.0 & 93.0 & 68.8 & 65.1 & 72.0  & 57.6 & 63.6 & 69.9 \\ 
\cite{fang2018learning}     & \textbf{72.8}      & \textbf{88.6 }      & \textbf{60.3 }  & \textbf{57.7}  & 62.7  & \textbf{47.5}  & \textbf{50.6}  & \textbf{60.4}     \\ \midrule 
Ours                 & 83.8     & 115.4      & 62.7   & 64.4  & 78.1  & 48.0  & 53.1  &  66.4    \\ \bottomrule
\end{tabular}}

\caption{ Mean per joint position error (MPJPE) in mm on the Human3.6M dataset.}
\label{tab:h36m_res}

\end{table}

Table \ref{tab:h36m_res} provides results of our method on the Human 3.6M dataset.  Let us notice that all models that achieve high performance on Human3.6m are single-person models that take as input cropped images containing a single fully visible subject. This setting is not representative of real world images where people can be anywhere in the image,  at various scales, truncated and occulted by other people.  The proposed model treats this general case and produces reliable results in a single person setting with an MPJPE of 66.4 mm on the Human 3.6M dataset, better than most compared approaches. In particular, it has a lower error than \cite{mehta2017single} that also uses ORPM but differs in the architecture used and in the joint grouping method.

\subsection{Multi-person 3D pose estimation on CMU-Panoptic}

CMU Panoptic \cite{Joo_2017_TPAMI} is a dataset containing images with several people performing different scenarios (playing an instrument, dancing, etc.) in a dome where several cameras are setup.  This dataset is challenging because of complex interactions and difficult camera viewpoints. We evaluate our model following these protocols: 

\begin{itemize}
    \item Panoptic-1 protocol: it is the protocol used in \cite{zanfir2018monocular,zanfir2018deep}. The model is evaluated on 9600 frames from HD cameras 16 and 30 and for 4 scenarios: Haggling, Mafia, Ultimatum, Pizza. The model is trained on the other 28 HD cameras of this dataset.
    \item Panoptic-2 protocol: This protocol is an extension of the previous one. Instead of evaluating on a subset of arbitrary selected frames, we evaluate on the entire sequences from cameras 16 and 30.  The training dataset in this protocol is the frames from all the HD cameras (except cameras 16 and 30) for the Haggling, Mafia, Ultimatum, Pizza scenarios. The model is evaluated on the same scenarios by taking one frame every ten frames from HD cameras 16 and 30. 
    \item Panoptic-3 protocol: Previous protocols use a large number of training cameras. To evaluate the robustness to the number of cameras and to the amount of training data, we propose protocol Panoptic-3. The model is trained on the Haggling, Mafia, Ultimatum, Pizza scenarios but only a subset of the training cameras is used: 
    \begin{itemize}
       
        \item Panoptic 3a: HD cameras 0, 2, 4, 6, 8, 10, 12, 14, 18, 20, 22, 24, 26 and 28 are used during training
         \item Panoptic 3b: HD cameras 0,4,8,12,20,24 and 28 are used during training
        \item Panoptic 3c: HD cameras 0,8, and 24 are used during training

    \end{itemize}
    The test set is the same as Panoptic 2. 
    \item Panoptic-4 protocol :  In the previous protocols, the model is trained and evaluated on the same scenarios. To evaluate the robustness to an unseen scenario in new camera viewpoints, we propose the Panoptic-4 protocol.   The training dataset in this protocol is the frames from all the HD cameras (except cameras 16 and 30) from the Haggling, Mafia and Ultimatum scenarios. The model is evaluated on the pizza scenario by taking one frame every ten frames from HD cameras 16 and 30. 
\end{itemize}

\textbf{Comparison with prior work}:   On Panoptic-1 protocol, our model improves the results over the recent state of the art methods on all the scenarios (Table \ref{tab:res_panoptic1}). It shows a global improvement of 5.0\% compared to \cite{zanfir2018deep}.  Note that unlike \cite{zanfir2018deep} we do not learn on any frame from the cameras 16 and 30 and on any external data. Actually, the proposed model does not need a trained attention readout process thanks to the effective ORPM readout process.

\textbf{Ablative studies: }  Table \ref{tab:res_panoptic1_ablative} provides ablative results of our method following  Panoptic-1 protocol on the Haggling, Mafia, Ultimatum and Pizza scenarios. Firstly, we present the results obtained by stacking one, two or three hourglass modules. Each time an hourglass module is added, the Mean per Joint Position Error (MPJPE) decreases (from 91.8 mm for one hourglass module to 68.5 mm for our full four hourglass modules model). This shows the importance of the stacking scheme and the refinement process in the model architecture.  The penultimate line of this table shows the results obtained with four hourglass modules and a Naive Readout (NR) in the ORPM, that means when the 3D joint coordinates are read directly from their 2D positions. Because of frequent crops and occlusions in the panoptic dataset, this model has poor performance with an MPJPE of 118.8 mm. This proves the importance of the ORPM storage redundancy to manage occlusion. Our complete model(last row) with four hourglass modules and the readout procedure described in Section \ref{sec:orpm} has the lowest MPJPE (68.5mm)

\begin{figure}

\begin{subfigure}{.9\textwidth}
  \centering
    \includegraphics[width=.9\linewidth, trim={0 13cm 0 13cm},clip]{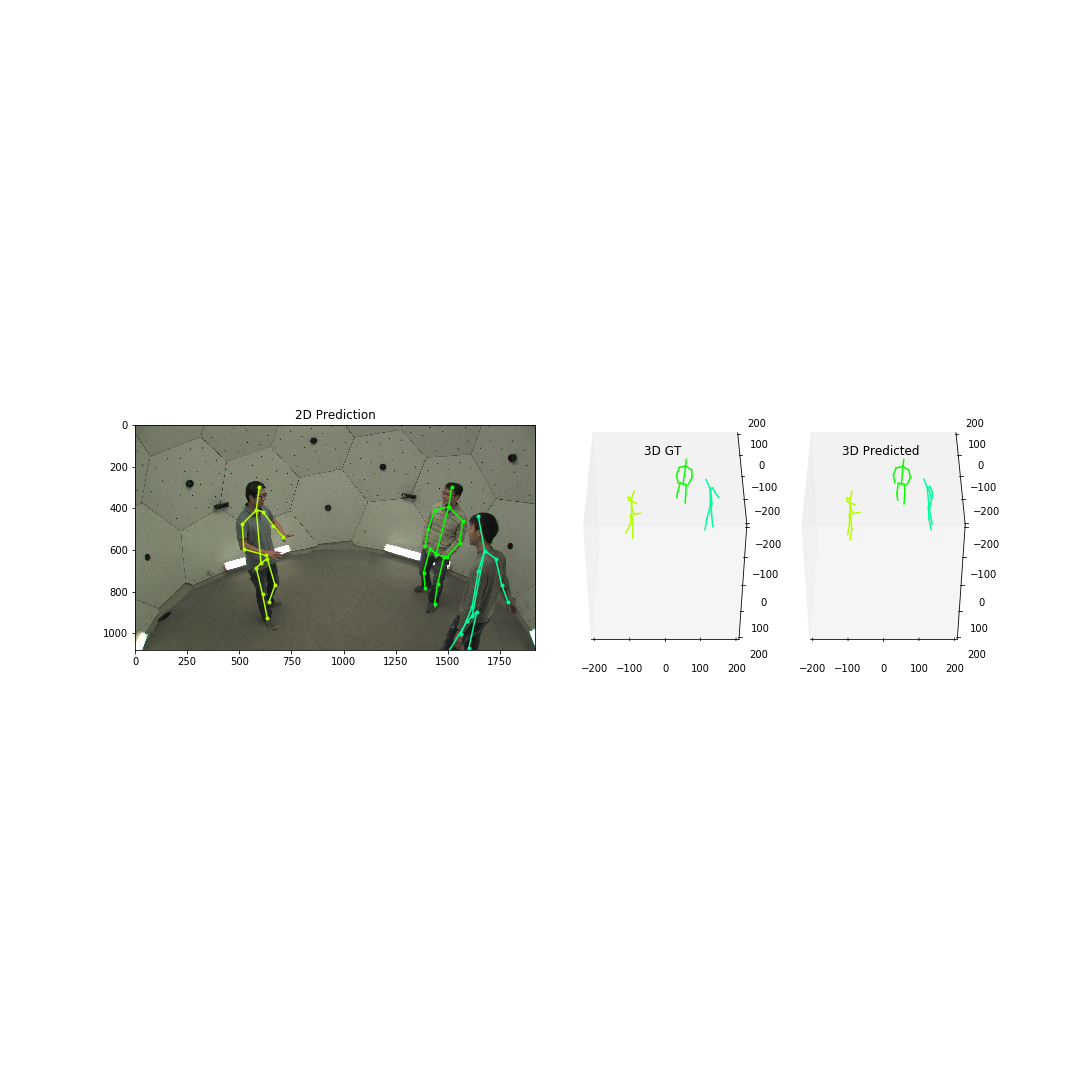}
\end{subfigure}

\begin{subfigure}{.9\textwidth}
  \centering
    \includegraphics[width=.9\linewidth, trim={0 13cm 0 13cm},clip]{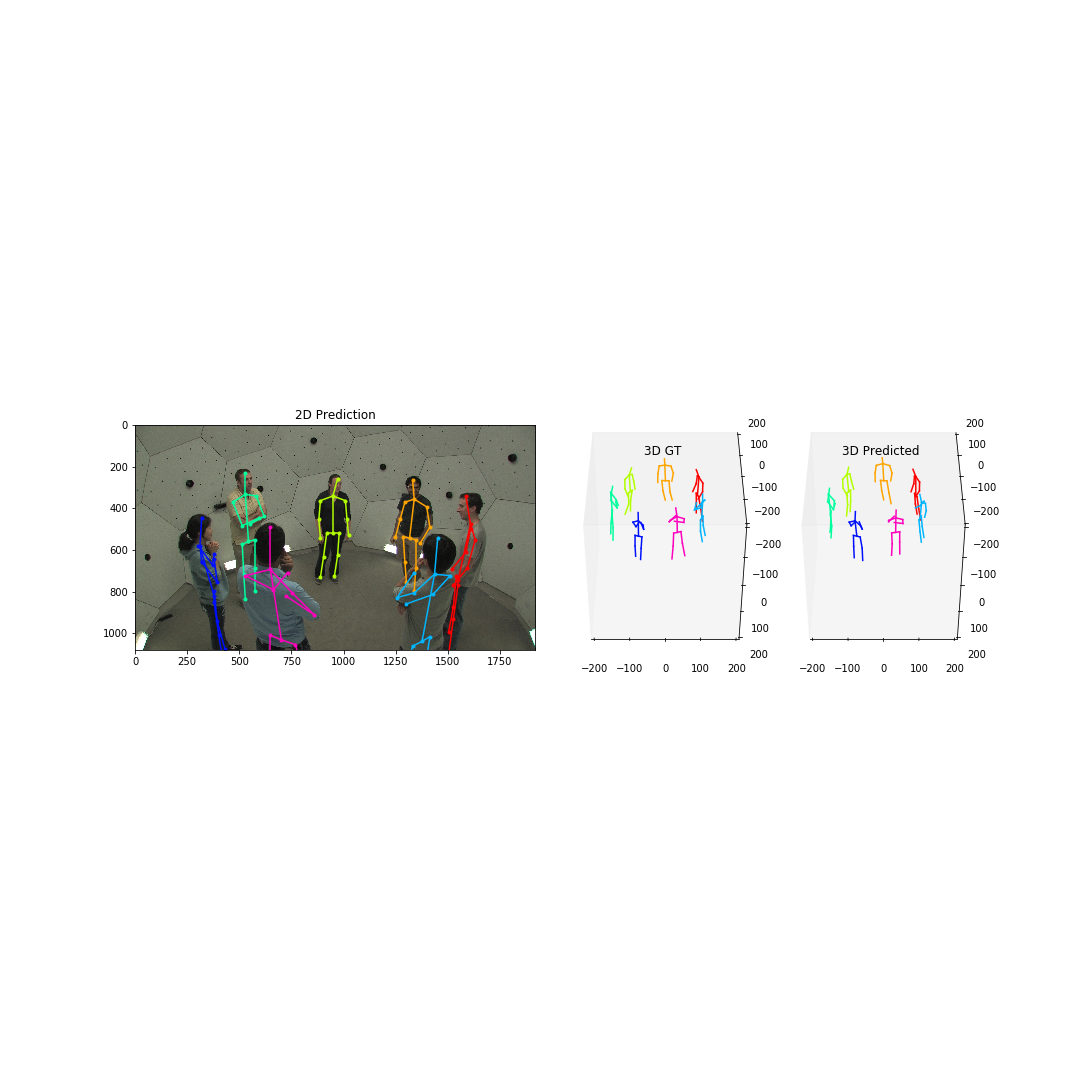}
\end{subfigure}

\begin{subfigure}{.9\textwidth}
  \centering
    \includegraphics[width=.9\linewidth, trim={0 13cm 0 13cm},clip]{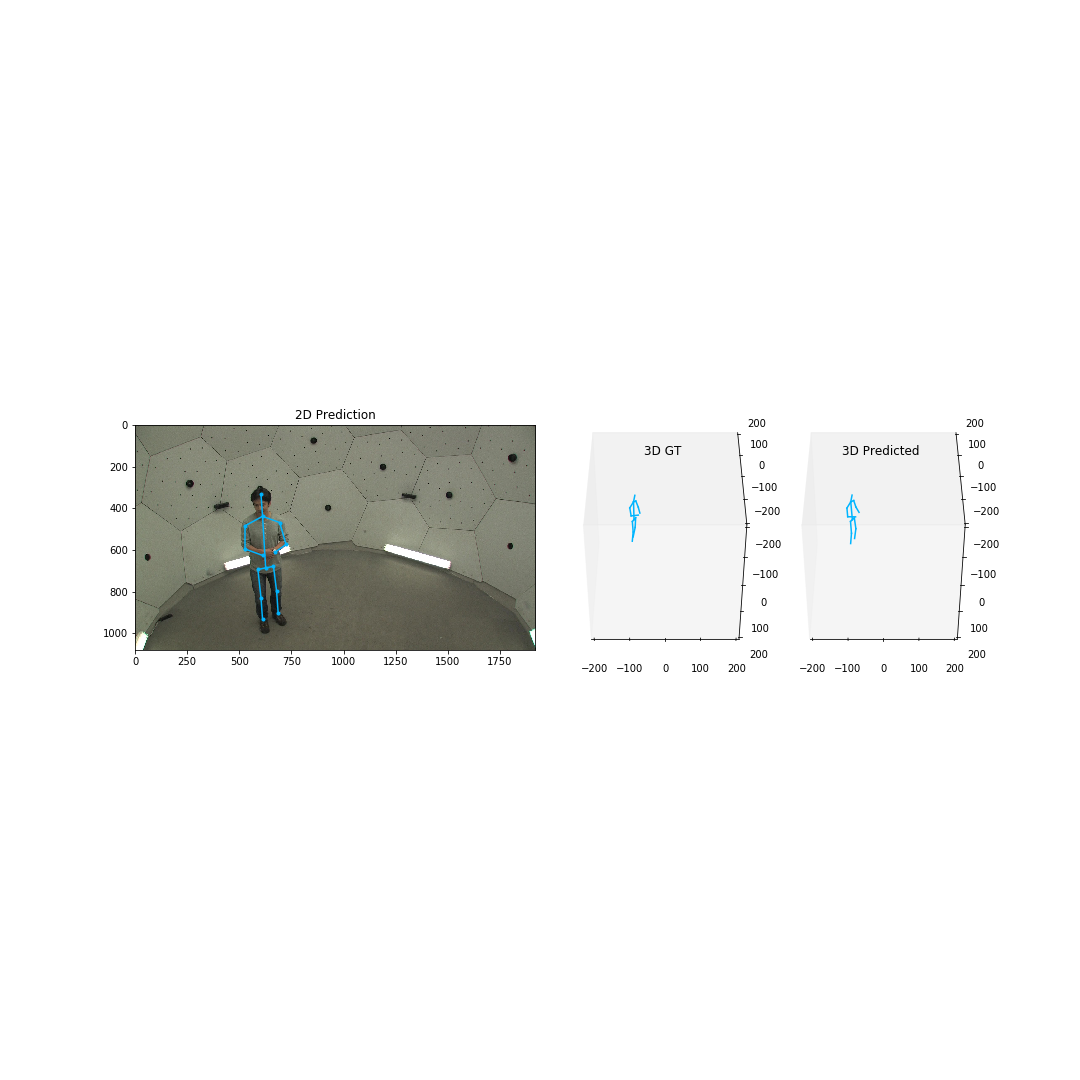}
\end{subfigure}

\begin{subfigure}{.9\textwidth}
  \centering
    \includegraphics[width=.9\linewidth, trim={0 13cm 0 13cm},clip]{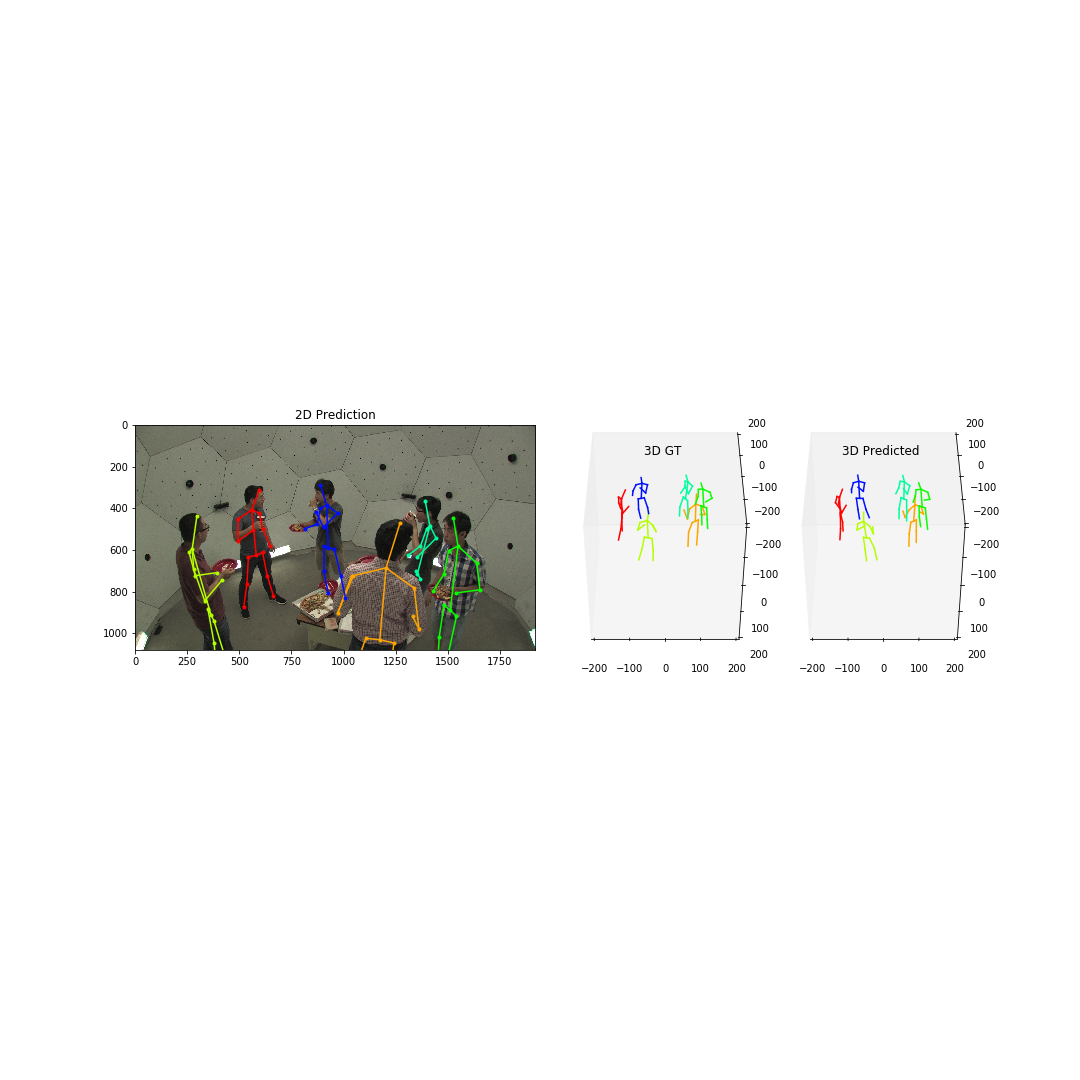}
\end{subfigure}

\caption{ Multi-person poses predicted by our approach on the CMU-Panoptic Dataset. Ground truth translation and scale are used for visualisation.  The first column corresponds to the input image with the predicted 2D pose. The second column corresponds to the ground truth 3D poses and the last column to the predicted 3D poses. These examples show that our approach works with a variable number of people in the image and can predict the 3D coordinates of joints that are not visible in the image thanks to the ORPM redundancy. Best viewed in color. }
\label{fig:res_panoptic}
\end{figure}

Examples of 3D human pose estimations on the Panoptic dataset are shown in Figure \ref{fig:res_panoptic}. Our method can estimate the 3D pose of multiple people even in case of truncation (1st, 2nd and last rows) or people overlap (2nd and 4th rows) 

\begin{table}[!t]
\centering
\resizebox{0.7\linewidth}{!}{%
\begin{tabular}{@{}llllll@{}}
\toprule
Method    & Haggling & Mafia & Ultimatum & Pizza & Mean  \\ \midrule
\cite{popa2017deep} & 217.9    & 187.3 & 193.6     & 221.3 & 203.4 \\
\cite{zanfir2018monocular}   & 140.0    & 165.9 & 150.7     & 156.0 & 153.4 \\
\cite{zanfir2018deep}   & 72.4     & 78.8  & 66.8      & 94.3  & 72.1  \\ \midrule
Ours, full       & \textbf{70.1}    & \textbf{66.6} & \textbf{55.6}     & \textbf{78.4}  & \textbf{68.5}  \\ \bottomrule
\end{tabular}
}
\caption{Mean per joint position error (MPJPE) in mm on the Panoptic Dataset following Panotic-1 Protocol}
\label{tab:res_panoptic1}
\end{table}

\begin{table}[!t]
\centering
\resizebox{\linewidth}{!}{%
\begin{tabular}{@{}l|ll|llll|l@{}}
\toprule
Method    & Nb of HG & ORPM & Haggling & Mafia & Ultimatum & Pizza & Mean  \\ \hline
Ours, 1-HG & 1       & \checkmark & 92.3     & 86.1 & 82.7     & 103.8 & 91.8\\
Ours, 2-HG & 2     & \checkmark & 77.1    & 74.8 & 68.0     &   89.8 & 78.3  \\
Ours, 3-HG & 3       & \checkmark & 72.4    & 72.4 &  60.12   & 85.2  & 73.8  \\
Ours, NR & 4        & $\times$ & 101.5    & 124.2 &  105.7   & 130.3  & 118.8  \\
Ours, full & 4      & \checkmark &\textbf{70.1}    & \textbf{66.6} & \textbf{55.6}     & \textbf{78.4}  & \textbf{68.5}  \\ \bottomrule
\end{tabular}
}
\caption{Mean per joint position error (MPJPE) in mm on the Panoptic Dataset following Panoptic-1 protocol. ($i$-HG stands for $i$ stacked hourglasses and NR for Naive Readout).}
\label{tab:res_panoptic1_ablative}
\end{table}

\par
\textbf{Robustness to the number of training cameras }: Protocols Panoptic 1 and 2 results are obtained by using a large number of training cameras.  What is the robustness of our model when using a reduced number of cameras ?  Table \ref{tab:res_panoptic_2_3_4} provides Panoptic 3 protocol results. Panoptic 3a and 3b results show that even by using only half and fourth of the training cameras, the MPJPE is only increased respectively by 6.9\% and 12.2\%. On the other hand, where only 3 training cameras are used, the MPJPE is 2.2 times greater than the Panoptic 2 MPJPE.  This number of cameras is insufficient to learn such a complex task. Even single person 3D human pose models are trained on datasets\cite{h36m_pami, mehta2017monocular} that provides images from four cameras or more.  

\par

\textbf{Performance on an useen scenario: } Protocols Panoptic 1,2 and 3 show the ability of the model to generalise to unseen camera viewpoints.  Panoptic 4 results show the ability of the model to generalise to new scenarios. The model is trained only on the Haggling, Mafia and Ultimatum scenarios and evaluated on the unseen Pizza scenario. The Panoptic 4 MPJPE (79.4) is close the MPJPE obtained on the Panoptic 2 protocol for the Pizza scenario showing that model does not overfeat on the training scenarios and can generalise to new ones.

\begin{table}[!t]
\centering
\resizebox{0.7\linewidth}{!}{%
\begin{tabular}{@{}llllll@{}}
\toprule
Protocol   & Haggling & Mafia & Ultimatum & Pizza & Mean  \\ \midrule
Panoptic 2 & 78.3   & 60.7 & 84.2   & 78.3 & 68.1 \\
Panoptic 3a & 82.4  & 64.3  & 88.7 & 82.2 & 72.3 \\
Panoptic 3b & 84.0   & 74.2 & 87.4  & 92.0 & 76.4 \\
Panoptic 3c & 149.4   & 151.3 & 155.5   & 167.9 & 150.9 \\
Panoptic 4 & \  & \ & \   & 79.4 & 79.4 \\

 \bottomrule
\end{tabular}
}
\caption{Mean per joint position error (MPJPE) in mm on the Panoptic Dataset following Panoptic-2, Panoptic-3 and Panoptic-4 protocols}
\label{tab:res_panoptic_2_3_4}
\end{table}

\par

\subsection{Multi-person 3D pose estimation on JTA dataset}

JTA (Joint Track Auto) is a dataset for human pose estimation and tracking in urban environment. It was collected from the realistic video-game the Grand Theft Auto V and contains 512 HD videos of 30 seconds recorded at 30 fps.  The collected videos feature a vast number of different body poses, in several urban scenarios
at varying illumination conditions and viewpoints. People perform different actions like walking, sitting, running, chatting, talking on the phone, drinking or smoking. Each image contains a number of people ranging between 0 and 60 with an average of more than 21 people. The distance from the camera ranges between 0.1 to 100 meters, resulting in pedestrian heights between 20 and 1100 pixels. None existing (virtual or real) dataset with annotated 3D pose is comparable with JTA dataset in terms of number of people per image, people and background variability.  As far as we know, we are the first to demonstrate the ability of a trained model to deal with such complex and rich environments with many people at different camera distances and with different resolutions. 256 videos are used for training and 128 for testing (the remaining 128 videos are used for validation). From the testing videos, we take one frame every ten frames for the evaluation. 

Table \ref{tab:res_jta} presents per camera distance results on the the JTA Dataset. We evaluate our model on this dataset at different resolutions (S1=512px, S2=1024px and S3=1536px) and also with the multi-scale inference described in section \ref{multi_scale_infrence}. The images from this dataset contain a large number of people in various distances from the camera. The distance from the camera can have a significant impact on the performance of a 3D human pose estimator. Indeed, distant people require higher image resolution and are more likely to be occulted. For this reason, we provide in Table \ref{tab:res_jta} results for people in different ranges of distance from the camera. Note that our testing set contains 262510 people. Among these people, 10\% have a distance from the camera less than 10 meters, 23\% have a distance from the camera between 10 and 20 meters, 21\% have a distance from the camera between 20 and 30 meters, 14\% have a distance from the camera between 30 and 40 meters and 31\% have a distance from the camera greater than 40 meters. 

The resolution having the best overall $3DPCK_r$ is the resolution S2 with a $3DPCK_r$ of 37.8\%. This resolution performs a good compromise to estimate the pose of the high resolution people that S3 cannot handle properly and low resolution people that are too small from scale S1.  Resolution S1 has the best results for people that are close to the camera (less than 10 meters) with an MPJPE of 165.2mm and a $3DPCK_r$ of 68.5\%. Resolution S2 has the best results for people that have a distance from the camera between 10 and 20 meters with a $3DPCK_r$ of 62.3\% and an MPJPE of 194.50. Resolution S3 has the best results for people that are far from the cameras (greater than 20 meters). These results show that each resolution is adequate to a given range of people distance and consequently to a resolution of people. 

The multi-scale inference (MSI) improves the overall $3DPCK_r$ and MPJPE. The $3DPCK_r$ goes from 37.8  to 43.9 for the MSI and the MPJPE goes from 258.9mm to 193.5mm. MSI has better results than scale S2 and S3 for  close to the camera people (less than 10 meters) taking advantages from poses estimated from scale S1 but without improving over this scale for these people.  It surpasses all the scales for people that have a distance from the camera greater than 10 meters. 

Joint-wise analysis (Table \ref{tab:res_jta_joint_wise}) shows that the results are unequal from one joint to another one. Regardless of the distance to the camera, spines and hips are always the best estimated joints. These articulations have a reduced variability compared to the extremity joints like wrists and ankles that have the worst MPJPE and $3DPCK_r$.  Indeed, since the 3D joint coordinates are expressed relatively to their parents joints in the kinematic tree and converted to pelvis relative locations,  errors in the estimation of a parent joint impact the estimation of all its descendent in the kinematic tree. One way to solve this issue could be to express the joints' coordinates relatively to more stable joints than their parent joints. For instance, the coordinates of the wrist could be expressed relatively to the elbow but also to the shoulder which is more stable and less prone to errors. A mechanism would be necessary to fuse these predictions expressed relatively to different joints and chose the more precise one. We leave this for a future work.

\begin{figure}

\begin{subfigure}{.9\textwidth}
  \centering
    \includegraphics[width=.4\linewidth, trim={5cm 11cm 5cm 10cm},clip]{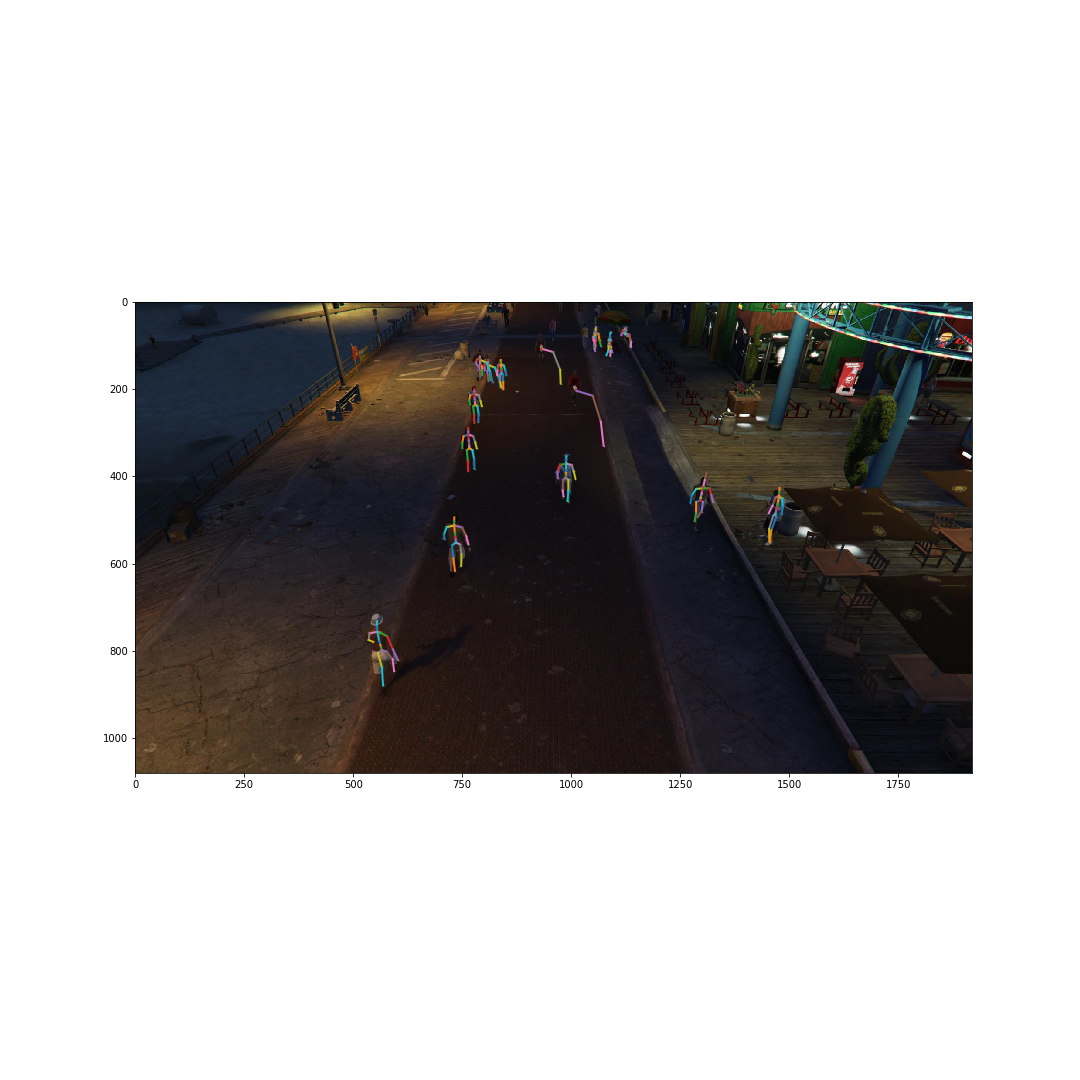}
        \includegraphics[width=.4\linewidth, trim={5cm 11cm 5cm 10cm},clip]{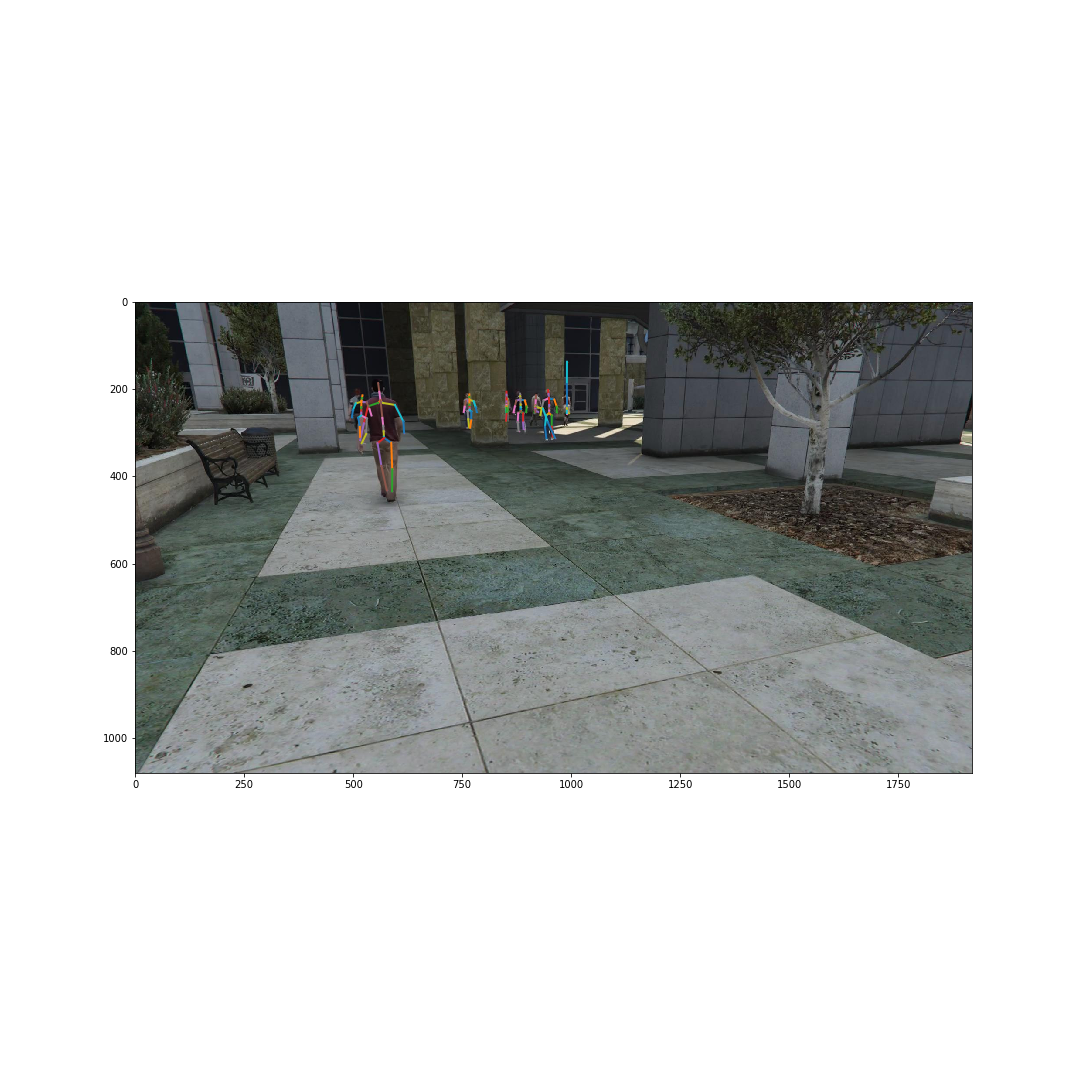}

\end{subfigure}

\begin{subfigure}{.9\textwidth}
  \centering
    \includegraphics[width=.4\linewidth, trim={5cm 11cm 5cm 10cm},clip]{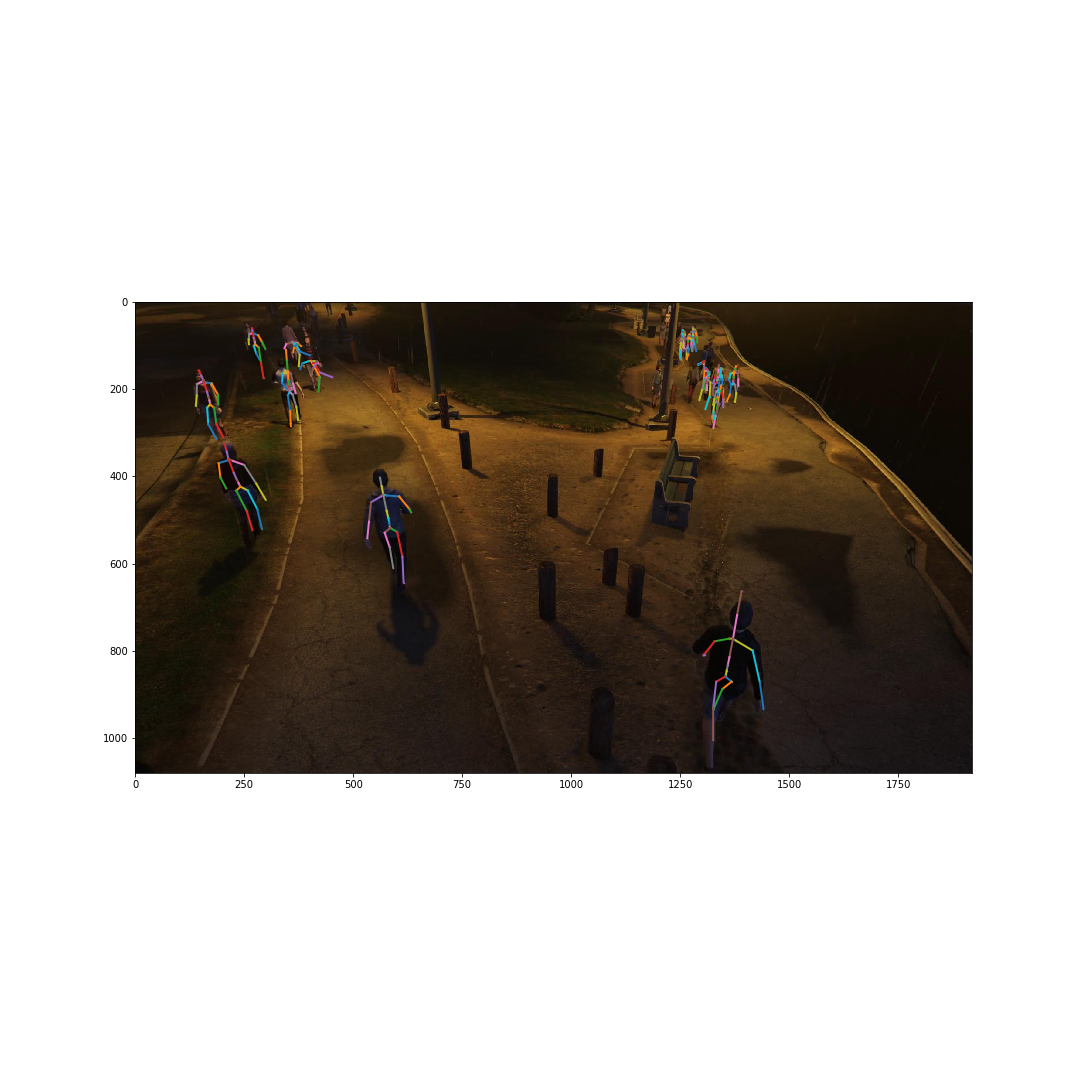}
    \includegraphics[width=.4\linewidth, trim={5cm 11cm 5cm 10cm},clip]{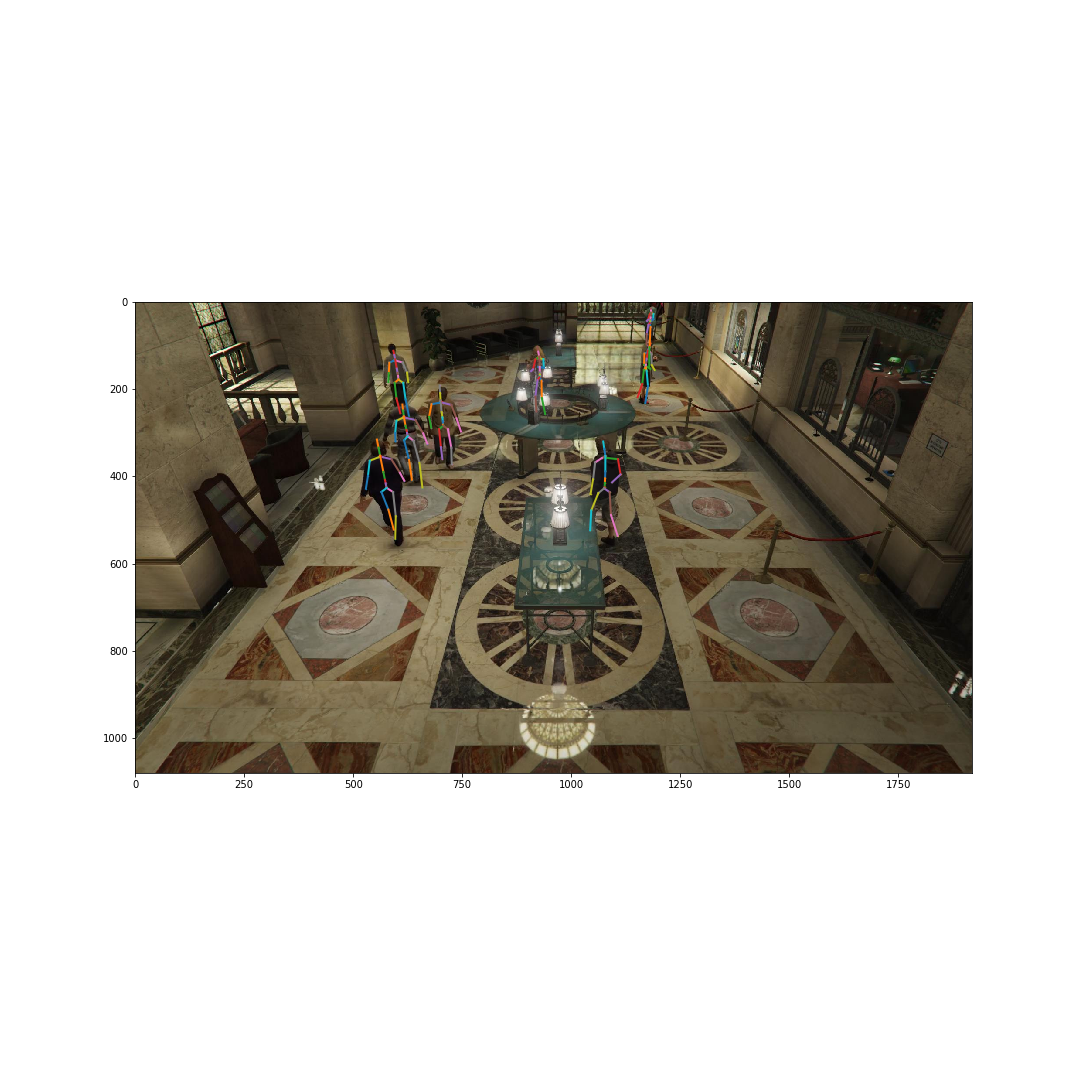}
\end{subfigure}

\begin{subfigure}{.9\textwidth}
  \centering
    \includegraphics[width=.4\linewidth, trim={5cm 11cm 5cm 10cm},clip]{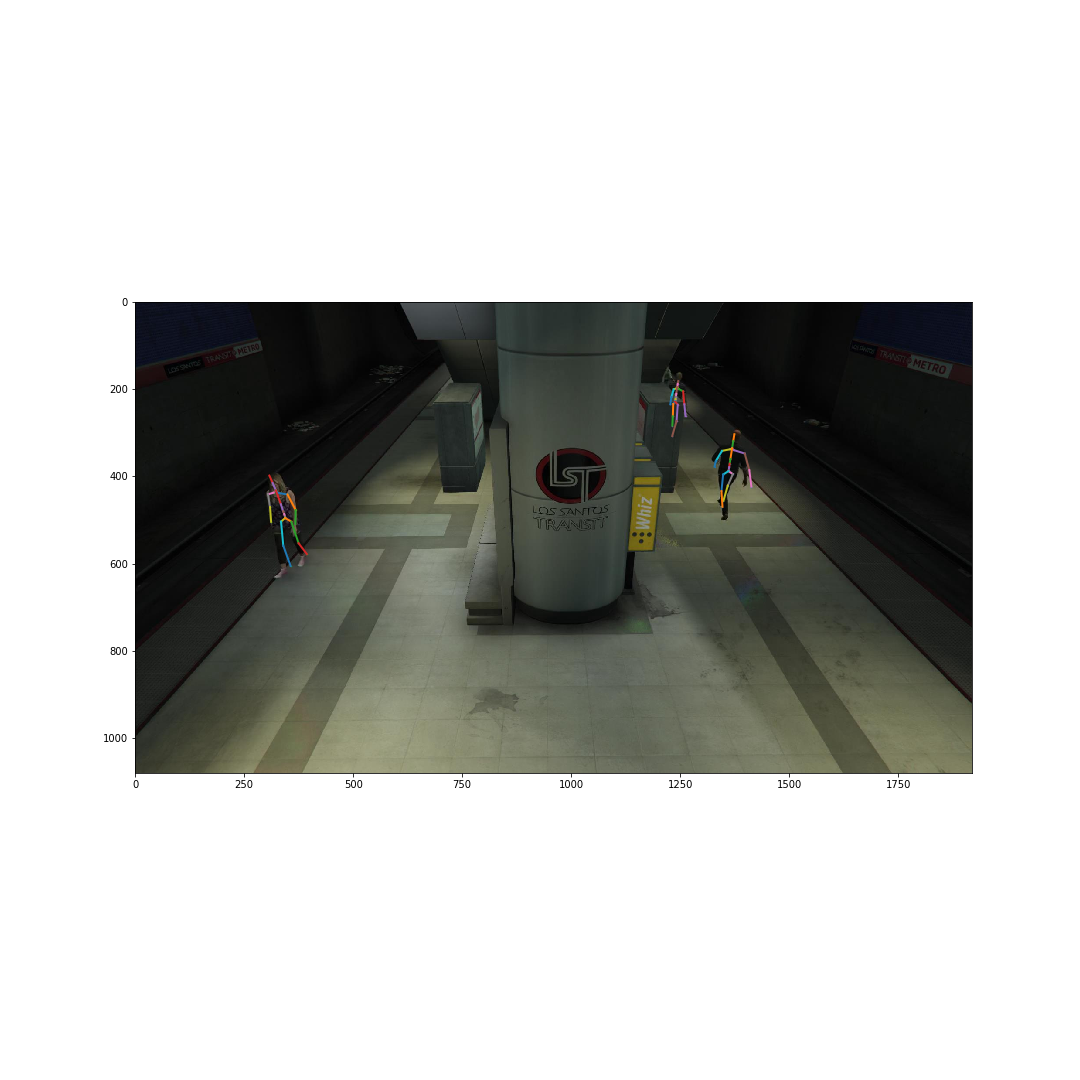}
    \includegraphics[width=.4\linewidth, trim={5cm 11cm 5cm 10cm},clip]{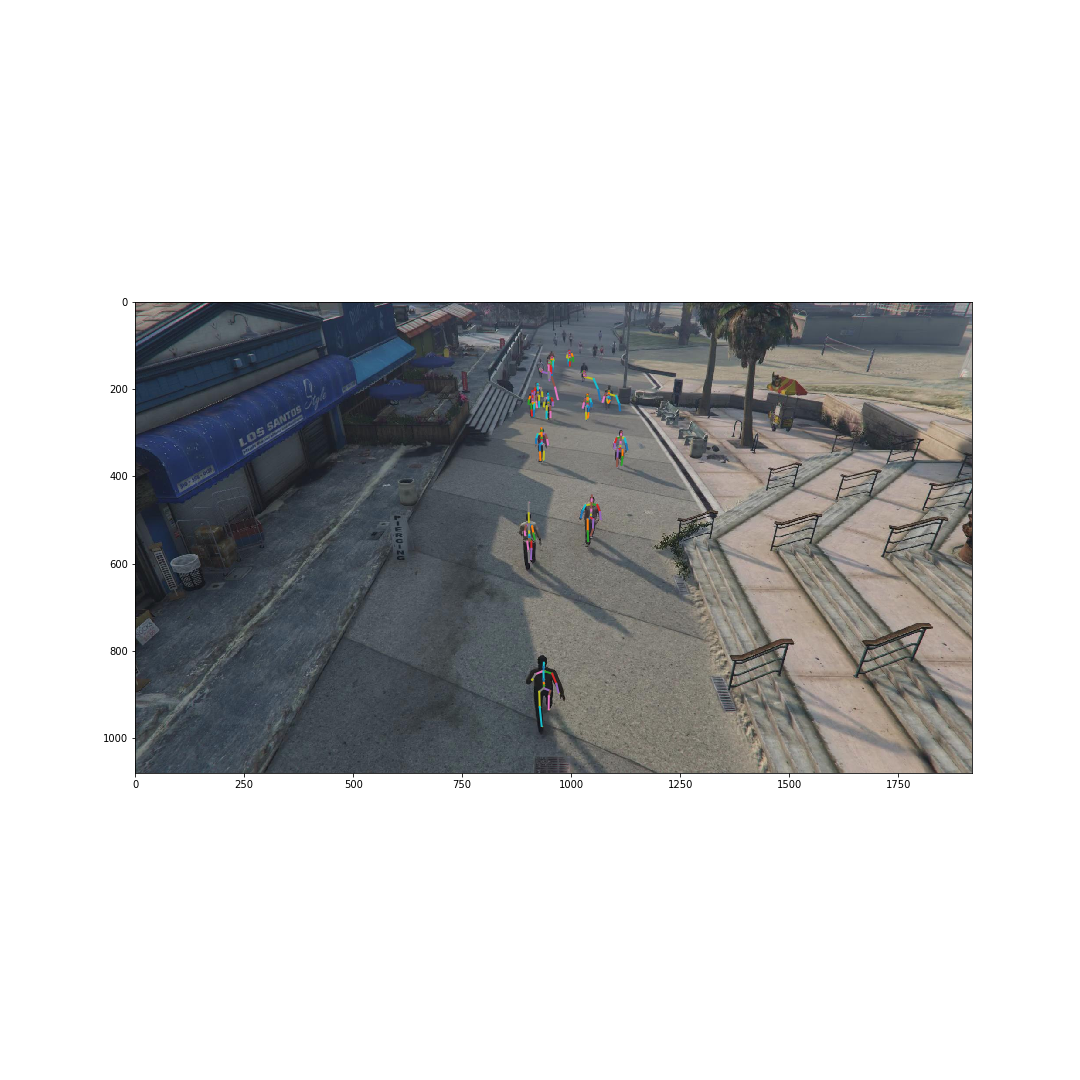}
\end{subfigure}


\caption{ Qualitative results (reprojected 3D poses) of our approach shown on the test set of JTA Dataset.  Our 3D estimations are relative to the pelvis. Ground truth translation and scale are used for visualisation. Our approach predict 3D poses in several urban scenarios at varying illumination conditions and viewpoints and for low resolution people. Nevertheless, very far people are not detected and the method fails in case of crowded people. Best viewed in color.}
\label{fig:res_jta}
\end{figure}

Examples of 3D human pose estimations on the JTA dataset are shown in Figure \ref{fig:res_jta}. Our method can estimate the 3D pose in several urban scenarios at varying illumination conditions and viewpoints.  Nevertheless, very far people are not detected and the method fails in case of crowded people. 

\begin{table}[]
\centering
\resizebox{0.9\textwidth}{!}{
\begin{tabular}{V{3}l V{3}l|l|l V{3}}
\Xhline{3\arrayrulewidth}
\textbf{Scale }              & \textbf{Distance to camera  } & \textbf{MPJPE} & \textbf{$3DPCK_r$ }   \\ \Xhline{3\arrayrulewidth}
\multirow{6}{*}{\textbf{S1 (512 px)}}   
                         & \multirow{1}{*}{\textless{}10m}  & 165.2 &  68.5   \\ \cline{2-4} 
                         & \multirow{1}{*}{\textgreater{}10m and \textless{}20m}  & 220.6 &  61.6   \\ \cline{2-4} 
                         & \multirow{1}{*}{\textgreater{}20m and \textless{}30m}    & 358.7  &  42.2   \\ \cline{2-4} 
                         & \multirow{1}{*}{\textgreater{}30m and \textless{}40m} &  409.7   &  36.0   \\ \cline{2-4} 
                         & \multirow{1}{*}{\textgreater{}40m}                    & 382.1   & 32.2   \\ \cline{2-4}
                         & \multirow{1}{*}{\textbf{All}} & \textbf{294.0}  & \textbf{33.1} \\ 
                         \Xhline{3\arrayrulewidth}
\multirow{6}{*}{\textbf{S2(1024px)}} 
                         & \multirow{1}{*}{\textless{}10 m}  &  275.53 & 43.5   \\ \cline{2-4} 
                         & \multirow{1}{*}{\textgreater{}10m and \textless{}20m}   &  194.50  &  62.3   \\ \cline{2-4} 
                         & \multirow{1}{*}{\textgreater{}20m and \textless{}30m}  &  281.5 &  51.25  \\ \cline{2-4} 
                         & \multirow{1}{*}{\textgreater{}30m and \textless{}40m} &  358.8  &  41.0   \\ \cline{2-4} 
                         & \multirow{1}{*}{\textgreater{}40m}  &  368.2  &  35.5   \\  \cline{2-4}
                         & \multirow{1}{*}{\textbf{All}}  & \textbf{ 258.9} &  \textbf{37.8}   \\ 
                         \Xhline{3\arrayrulewidth}
\multirow{6}{*}{\textbf{S3(1536px)}}   
                         & \multirow{1}{*}{\textless{}10m}  &  319.0  & 33.9   \\ \cline{2-4} 
                         & \multirow{1}{*}{\textgreater{}10m and \textless{}20m}  &  231.16 & 49.4   \\ \cline{2-4} 
                         & \multirow{1}{*}{\textgreater{}20m and \textless{}30m} &  222.75 & 53.3   \\ \cline{2-4} 
                         & \multirow{1}{*}{\textgreater{}30m and \textless{}40m} &   269.1 &  47.5   \\ \cline{2-4} 
                         & \multirow{1}{*}{\textgreater{}40m} &  305.90  &  38.8   \\ \cline{2-4}
                          & \multirow{1}{*}{\textbf{All}}   &  \textbf{274.3} &  \textbf{34.8}   \\ 
                         \Xhline{3\arrayrulewidth}

\multirow{6}{*}{\textbf{Multi Scale Inference(MSI)}}  
                         & \multirow{1}{*}{\textless{}10m}  & 175.5  & 55.8   \\ \cline{2-4} 
                         & \multirow{1}{*}{\textgreater{}10m and \textless{}20m} & 137.3 &  68.4   \\ \cline{2-4} 
                         & \multirow{1}{*}{\textgreater{}20m and \textless{}30m}   &  187.3  &  57.8  \\ \cline{2-4} 
                         & \multirow{1}{*}{\textgreater{}30m and \textless{}40m}  &  231.8  &  49.3   \\ \cline{2-4} 
                         & \multirow{1}{*}{\textgreater{}40m} & 262.1 &  41.7   \\ \cline{2-4} 
                          & \multirow{1}{*}{\textbf{All}} & \textbf{193.5} &  \textbf{43.9}   \\ 
\Xhline{3\arrayrulewidth}
\end{tabular}
}
\caption{MPJPE and $3DPCK_r$ on the JTA dataset. Results are provided per scale and per camera distance that means by taking into account in the metrics computation only the people that are in the corresponding distance range from the camera.}
\label{tab:res_jta}
\end{table}

\begin{table}[]
\centering
\resizebox{\textwidth}{!}{
\begin{tabular}{|l|l|l|l|l|l|l|l|l|l|l|l|l|}
\hline
 Distance to camera                                    & Metric & head  & neck  & clavicles & shoulders & elbows & wrists & spines & hips  & knees & ankles & all    \\ \hline
     
 \multirow{2}{*}{\textgreater{}0}                      & MPJPE  & 196.5 & 174.7 & 174.9     & 215.3     & 264.6  & 329.4 & 42.3   & 76.3  & 253.2 & 425.5  & 193.5  \\ 
                                                                                & $3DPCK_r$ & 41.1  & 44.6 &   44.9      & 33.8      & 27.2   & 19.0   & 74.4  & 73.9  & 25.7  & 8.9    & 43.9   \\ \hline
                          \multirow{2}{*}{\textless{}10m}                       & MPJPE  & 131.7 & 195.1 & 191.8    & 219.5    & 218.7  & 254.6  & 45.8   & 66.97 & 236.1 & 395.9 & 175.5  \\  
                                                     & $3DPCK_r$ & 68.1  & 48.1  & 48.5      & 37.5       & 39.5    & 30.6    & 94.2  & 94.0  & 29.0   & 7.3    & 55.8   \\ \hline
                          \multirow{2}{*}{\textgreater{}10m and \textless{}20m} & MPJPE  & 117.4 & 115.4 & 117.5     & 152.9     & 186.9  & 237.5  & 29.8   & 60.2  & 189.0 & 317.0  & 137.3 \\ 
                                                                               & $3D_PCK$ & 76.2  & 115.4  & 75.9      & 62.9      & 55.2   & 40.4   & 98.5   & 93.1  & 46.8  & 17.8   & 68.4   \\ \hline 
                          \multirow{2}{*}{\textgreater{}20m and \textless{}30m} & MPJPE  & 162.3 & 133.0 & 140.4     & 200.2   & 270.0  & 348.1  & 34.1   & 85.1  & 264.9 & 437.7  & 187.3 \\ 
                                                                               & $3DPCK_r$ & 61.0 & 70.6  & 67.8      & 46.5     & 33.5  & 20.8   & 97.5   & 85.8  & 30.2  & 11.1   & 57.8  \\ \hline
                          \multirow{2}{*}{\textgreater{}30m and \textless{}40m} & MPJPE  & 211.4 & 166.4 & 177.6     & 257.6     & 347.3  & 347.3  & 41.2   & 105.1 & 311.9 & 516.9  & 231.8  \\ 
                                                                                & $3DPCK_r$ & 48.0  & 60.9  & 56.0      & 30.3      & 19.8   & 11.0  & 95.7  & 79.8  &20.8  & 6.8    & 49.3   \\ \hline
                          \multirow{2}{*}{\textgreater{}40m}                    & MPJPE  & 248.0 & 200.3 & 212.1     & 310.8    & 410.6  & 505.2  & 49.7   & 119.0 & 324.7 & 528.5 & 262.1 \\  
                                                                               & $3DPCK_r$ & 39.2  & 50.1  & 45.3      & 18.0      & 11.1   & 6.0    & 89.7   & 72.1  & 13.1  & 4.5    & 41.7   \\ \hline
\end{tabular}
}

\caption{Joint wise MPJPE and $3DPCK_r$ on the JTA Dataset of our approach with the Multi-Scale Inference}
\label{tab:res_jta_joint_wise}
\end{table}

\subsection{Multi-person 3D pose estimation on MuPoTS-3D dataset}

MuPoTS-3D  \cite{mehta2017single} is a dataset containing 20 indoor and outdoor sequences with ground truth 3D poses for up to three subjects.  Like Mehta \textit{et. al} \cite{mehta2017single},  our model is trained on the MuCo-3DHP dataset that has been generated by compositing the existing MPI-INF-3DHP 3D single-person pose estimation dataset \cite{mehta2016monocular} and the COCO-dataset \cite{lin2014microsoft} to ensure better generalisation. Each mini-batch consists of half MuCo-3DHP and half COCO images. For COCO data, the loss value for the ORPM is set to zero.

We compare our approach with the single-shot approach proposed by \citet{mehta2017single} and a recent two-stage approach \cite{moon2019camera}.  Like \cite{mehta2017single}, our model is based on the ORPM formulation but differs in the stacked architecture used and in the bottom-up joints association method.  Table  \ref{tab:mupots} provides  $3DPCK_r$  results on this dataset. Our model achieves higher accuracy  with a $3DPCK_r$ of 67.5\% ( 72.7\% when evaluating only on well detected people) compared to the approach of Mehta \textit{et. al} \cite{mehta2017single} that has a $3DPCK_r$ of 65\% (69.8 for matched people). 
Compared to the approach of Moon \textit{et. al.} \cite{moon2019camera}, our approach has a lower $3DPCK_r$. This top-down approach uses Faster-R CNN \cite{ren2015faster}, an external two-stage object detector to compute human bounding boxes. Each cropped box is then forwarded to a single-person 3D person approach \cite{sun2018integral}. Consequently, the computational complexity of this model depends on the number of people in the image. If this number is large, this approach scales badly while the proposed bottom-up model has a constant inference time regarding the number of people. 

 While absolute 3D human pose estimation is not in the scope of this work, absolute 3D poses can be obtained by minimising the reprojection error between  predicted root-relative 3D poses and 2D poses. Table  \ref{tab:mupots_abs} provides  $3DPCK_a$  results on the MuPoTS dataset. Our simple baseline can be improved by more sophisticated methods that take into account image features, joints occlusions, ground plane information and temporal contexts. These improvements will be studied in a future work.

  \begin{landscape}
\begin{table}[]
\resizebox{\linewidth}{!}{

\begin{tabular}{ll|lllllllllllllllllllll}
\multicolumn{2}{l}{All}                                                                     & S1                             & S2                             & S3                             & S4                             & S5                             & S6                             & S7                             & S8                             & S9                             & S10                             & S11                            & S12                            & S13                            & S14                             & S15                            & S16                            & S17                            & S18                            & S19                            & \multicolumn{1}{l|}{S20}                            & Avg                            \\ \hline
\multicolumn{1}{l|}{Two-Stage}                    & \cite{moon2019camera}            & 94.4                           & 77.5                           & 79.0                           & 81.9                           & 85.3                           & 72.8                           & 81.9                           & 75.7                           & 90.2                           & 90.4                            & 79.2                           & 79.9                           & 75.1                           & 72.7                            & 81.1                           & 89.9                           & 89.6                           & 81.8                           & 81.7                           & 76.2                                                & 81.8                           \\ \hline
\multicolumn{1}{l|}{\multirow{2}{*}{Single-Shot}} & \cite{mehta2017single} & \textbf{81.0} & 59.9                           & \textbf{64.4} & 62.8                           & 68.0                           & 30.3                           & 65.0                           & 59.2                           & \textbf{64.1} & \textbf{ 83.9} & 67.2                           & 68.3                           & 60.6                           & \textbf{56.5} & \textbf{69.9} & 79.4                           & \textbf{79.6} & 66.1                           & 64.3                           & \multicolumn{1}{l|}{63.5}                           & 65.0                           \\
\multicolumn{1}{l|}{}                             & Ours                                    & 78.1                           & \textbf{62.5} & 55.5                           & \textbf{63.8} & \textbf{70.2} & \textbf{50.8} & \textbf{73.8} & \textbf{65.3} & 55.1                           & 79.3                            & \textbf{70.4} & \textbf{72.3} & \textbf{65.4} & 55.3                            & 65.2                           & \textbf{81.3} & 77.2                           & \textbf{75.9} & \textbf{74.2} & \multicolumn{1}{l|}{\textbf{71.6}} & \textbf{67.5}
\end{tabular}
}

\resizebox{\linewidth}{!}{%
\begin{tabular}{ll|lllllllllllllllllllll}
\multicolumn{2}{c}{Matched}                                                & S1            & S2            & S3            & S4            & S5            & S6            & S7            & S8            & S9            & S10           & S11           & S12           & S13           & S14            & S15           & S16           & S17           & S18           & S19           & \multicolumn{1}{l|}{S20}            & Total         \\ \hline
\multicolumn{1}{l|}{Two-Stage}                    & \cite{moon2019camera}            & 94.4          & 78.6          & 79.0          & 82.1          & 86.6          & 72.8          & 81.9          & 75.8          & 90.2          & 90.4          & 79.4          & 79.9          & 75.3          & 81.0           & 81.0          & 90.7          & 89.6          & 83.1          & 81.7          & 77.3                                & 82.5          \\ \hline
\multicolumn{1}{l|}{\multirow{2}{*}{Single-Shot}} & \cite{mehta2017single} & \textbf{81.0} & 64.3          & \textbf{64.6} & 63.7          & 73.8          & 30.3          & 65.1          & 60.7          & \textbf{64.1} & \textbf{83.9} & 71.5          & 69.6          & 69            & \textbf{69.6 } & \textbf{71.1} & 82.9          & 79.6          & 72.2          & 76.2          & \multicolumn{1}{l|}{85.9}           & 69.8          \\
\multicolumn{1}{l|}{}                             & Ours                   & 78.3          & \textbf{75.0} & 56.9          & \textbf{64.1} & \textbf{76.1} & \textbf{51.3} & \textbf{74.7} & \textbf{79.1} & 55.2          & 79.3          & \textbf{74.5} & \textbf{74.5} & \textbf{70.2} & 69.5           & 67.6          & \textbf{85.7} & \textbf{82.6} & \textbf{78.7} & \textbf{79.1} & \multicolumn{1}{l|}{\textbf{ 89.6}} & \textbf{72.7}
\end{tabular}
}
\caption{Sequence-wise (TS1-TS10) $3DPCK_r$ of our method and other state of the art methods on the MuPoTS-3D dataset. We report both the overall $3DPCK_r$ (first two lines) and the $3DPCK_r$ only for people matched to a ground truth (last two lines)}
\label{tab:mupots}
\end{table}

\end{landscape}

  \begin{landscape}
\begin{table}[]
\resizebox{\linewidth}{!}{
\begin{tabular}{l|llllllllllllllllllll|l}
All         & S1   & S2   & S3   & S4   & S5   & S6   & S7   & S8   & S9   & S10  & S11  & S12  & S13  & S14  & S15  & S16  & S17  & S18  & S19  & S20  & Total \\ \hline
\cite{moon2019camera} & 59.5 & 44.7 & 51.4 & 46.0 & 52.2 & 27.4 & 23.7 & 26.4 & 39.1 & 23.6 & 18.3 & 14.9 & 38.2 & 26.5 & 36.8 & 23.4 & 14.4 & 19.7 & 18.8 & 25.1 & 31.5  \\
Ours        & 22.7 & 18.1 & 16.1 & 18.5 & 20.4 & 14.7 & 21.2 & 18.9 & 16.0 & 22.9 & 20.3 & 20.9 & 18.9 & 16.0 & 18.9 & 23.5 & 22.3 & 21.8 & 21.5 & 20.8 & 19.8 
\end{tabular}
}

\resizebox{\linewidth}{!}{%
\begin{tabular}{l|llllllllllllllllllll|l}
Matched     & S1   & S2   & S3   & S4   & S5   & S6   & S7   & S8   & S9   & S10  & S11  & S12  & S13  & S14  & S15  & S16  & S17  & S18  & S19  & S20  & Total \\ \hline
\cite{moon2019camera} & 59.5 & 45.3 & 51.4 & 46.2 & 53.0 & 27.4 & 23.7 & 26.4 & 39.1 & 23.6 & 18.3 & 14.9 & 38.2 & 29.5 & 36.8 & 23.6 & 14.4 & 20.0 & 18.8 & 25.4 & 31.5  \\
Ours        & 22.7 & 21.2 & 17.1 & 18.6 & 22.0 & 14.8 & 21.5 & 22.9 & 16.0 & 22.9 & 21.5 & 21.6 & 20.3 & 20.0 & 19.4 & 18.9 & 23.8 & 22.6 & 22.9 & 25.8 & 20.9 
\end{tabular}
}
\caption{Sequence-wise (TS1-TS10) $3DPCK_a$ of our method and a recent method of the state of the art on the MuPoTS-3D dataset. We report both the overall $3DPCK_a$ (first two lines) and the $3DPCK_a$ only for people matched to a ground truth (last two lines)}
\label{tab:mupots_abs}
\end{table}

\end{landscape}

 \begin{figure}

\begin{subfigure}{.9\textwidth}
  \centering
    \includegraphics[width=.9\linewidth,clip]{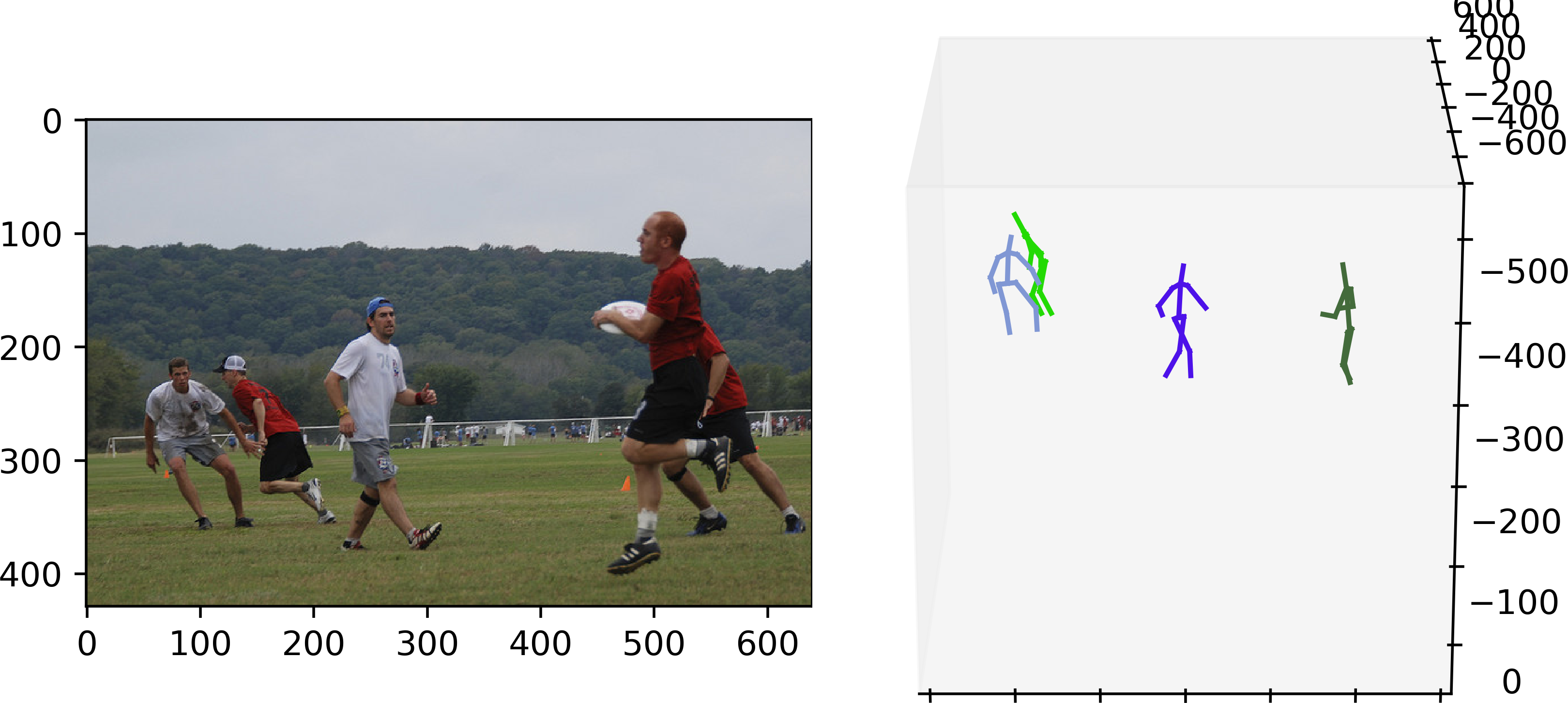}
\end{subfigure}

\begin{subfigure}{.9\textwidth}
  \centering
    \includegraphics[width=.9\linewidth, clip]{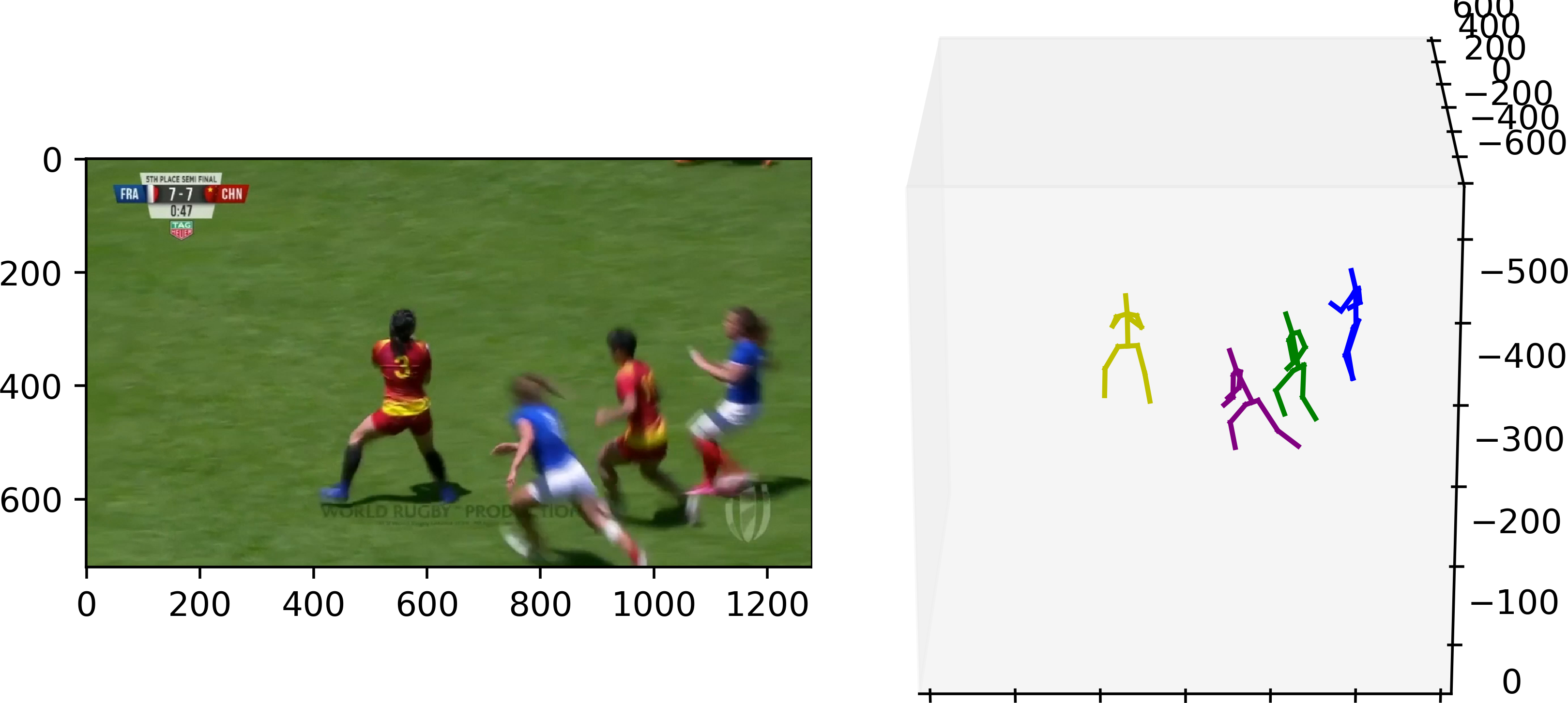}
\end{subfigure}

\begin{subfigure}{.9\textwidth}
  \centering
    \includegraphics[width=.9\linewidth,clip]{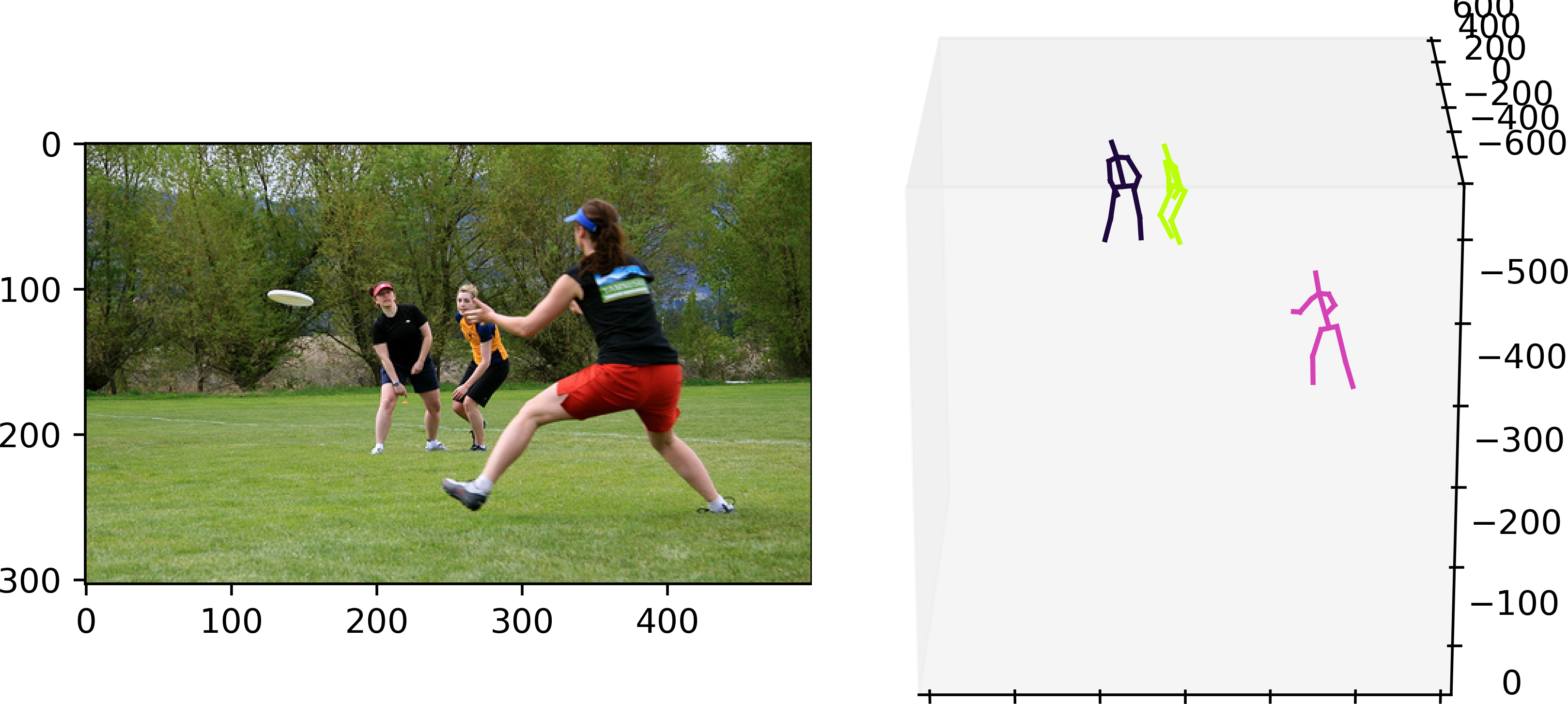}
\end{subfigure}

\caption{ Multi-person poses predicted by our approach on natural images. The first column corresponds to the input image. The second column corresponds to predicted 3D poses. Best viewed in color. }
\label{fig:res_itw}
\end{figure}

\section{Conclusion}

We have presented a single shot trainable model for multi-person 3D pose estimation
in real or virtual complex environments 
2D and 3D human joints are predicted using heatmaps and ORPM which have proven their ability to
manage occlusions. The difficult problem of associating joints to people skeletons is
managed using the recent associative embeddings method. Additionally, the same stacked network
jointly learns and estimates, in an end-to-end manner, 2D human poses and 3D human
poses exploiting the complementarity of these tasks.

The experiments provided in this work have proven the importance of the stacking scheme
and the ORMP formulation, validating the proposed network architecture. Furthermore, large-
scale experiments, on the CMU Panoptic dataset, demonstrate that the proposed approach
results surpass those of the state of the art. Experimental results on the MuPoTS-3D dataset
show the high accuracy of our model on both outdoor and indoor multi-person scenes.

Experiments on the JTA Dataset are accurate for people close to the camera, even in
crowed situation. Nevertheless, more complex urban scenarios involving many people at
different image resolution remain challenging. These experiments show also that the tree structure induces more error on the extremity joints. One
way to solve this problem is to express the joints’ coordinates relatively to more stable joints
than their parent joints. The choice of these stable joints and their use will be the subject of
future work.

\bibliographystyle{model1-num-names}
\bibliography{main.bib}







\end{document}